\documentclass[letterpaper]{article} % DO NOT CHANGE THIS
\usepackage{aaai25}  % DO NOT CHANGE THIS
\usepackage{times}  % DO NOT CHANGE THIS
\usepackage{helvet}  % DO NOT CHANGE THIS
\usepackage{courier}  % DO NOT CHANGE THIS
\usepackage[hyphens]{url}  % DO NOT CHANGE THIS
\usepackage{graphicx} % DO NOT CHANGE THIS
\usepackage{natbib}  % DO NOT CHANGE THIS AND DO NOT ADD ANY OPTIONS TO IT
\usepackage{caption} % DO NOT CHANGE THIS AND DO NOT ADD ANY OPTIONS TO IT
\usepackage{algorithm}
\usepackage{algorithmic}

\usepackage{multirow}
\usepackage{pifont}
\usepackage{amsmath}
\usepackage[linesnumbered,ruled,vlined,algo2e]{algorithm2e}
\usepackage{amssymb}
\usepackage{newfloat}
\usepackage{listings}
\DeclareCaptionStyle{ruled}{labelfont=normalfont,labelsep=colon,strut=off} % DO NOT CHANGE THIS
\floatstyle{ruled}
\newfloat{listing}{tb}{lst}{}
\floatname{listing}{Listing}
\title{Dual-Level Precision Edges Guided Multi-View Stereo \\ with Accurate Planarization}
\author{
    Kehua Chen,
    Zhenlong Yuan,
    Tianlu Mao\thanks{Corresponding author.},
    Zhaoqi Wang
}
\affiliations{
    Institute of Computing Technology, Chinese Academy of Sciences \\
    \{chenkehua23s,yuanzhenlong21b,ltm,zqwang\}@ict.ac.cn
}

\usepackage{bibentry}
\begin{document}

\maketitle

\begin{abstract}
The reconstruction of low-textured areas is a prominent research focus in multi-view stereo (MVS). In recent years, traditional MVS methods have performed exceptionally well in reconstructing low-textured areas by constructing plane models. However, these methods often encounter issues such as crossing object boundaries and limited perception ranges, which undermine the robustness of plane model construction. Building on previous work (APD-MVS), we propose the DPE-MVS method. By introducing dual-level precision edge information, including fine and coarse edges, we enhance the robustness of plane model construction, thereby improving reconstruction accuracy in low-textured areas. Furthermore, by leveraging edge information, we refine the sampling strategy in conventional PatchMatch MVS and propose an adaptive patch size adjustment approach to optimize matching cost calculation in both stochastic and low-textured areas. This additional use of edge information allows for more precise and robust matching. Our method achieves state-of-the-art performance on the ETH3D and Tanks \& Temples benchmarks. Notably, our method outperforms all published methods on the ETH3D benchmark.
\end{abstract}


\section{Introduction}

Multi-view stereo (MVS) is a classical computer vision task aimed at reconstructing the dense 3D geometry of objects or scenes from images taken from multiple viewpoints. This technique has significant applications in areas such as cultural heritage preservation, virtual reality, augmented reality, and autonomous driving. In recent years, MVS methods has advanced significantly, benefiting from diverse datasets \cite{Schops2017,Knapitsch2017} and various algorithms \cite{Wang2023,Wu2024} , leading to substantial improvements in reconstruction performance. Despite these advancements, MVS still faces challenges in handling low-textured and stochastic textured areas.

MVS methods can be roughly categorized into traditional methods \cite{Galliani2015,Schoenberger2016,Xu2019} and learning-based methods \cite{Yao2018,Gu2020}. Traditional methods have the advantages of stronger generalization capabilities and lower memory consumption compared to learning-based methods. Additionally, there has been more research in recent traditional MVS methods addressing the low-textured issue. Recent mainstream methods are based on PatchMatch (PM), which matches fixed-size patch in the reference image with patches in the source images using a plane hypothesis (including depth and normal). Since fixed-size patch struggle to extract appropriate feature information in low-textured areas, many works have further extended and optimized this method. For example, \cite{Liao2019,Xu2019} leverage multi-scale information, while others \cite{Xu2020} use triangular plane priors to guide plane hypotheses in low-textured areas. \cite{Xu2022} combines these approaches to enhance reconstruction performance. Subsequent methods \cite{Zhang2022, Tian2023} refine the construction of triangular planes, while others \cite{Romanoni2019, kuhn2019plane} employ image segmentation and RANSAC algorithm to determine plane models.

\begin{figure}[t]
\centering
\includegraphics[width=1\columnwidth]{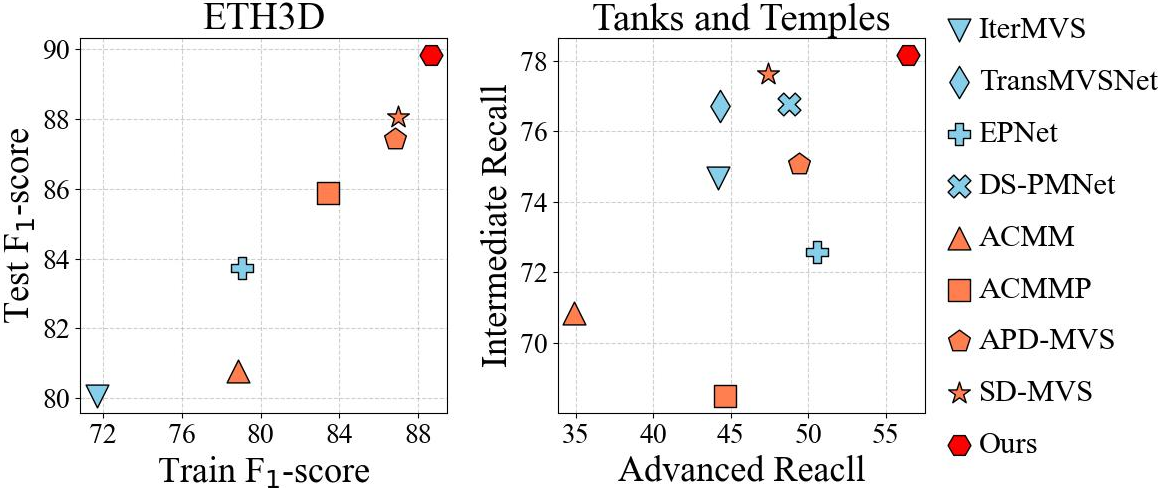
} % Reduce the figure size so that it is slightly narrower than the column. Don't use precise values for figure width.This setup will avoid overfull boxes.
\caption{Comparison with the SOTA traditional methods and learning-based methods. Our method achieves the best F$_1$-score on ETH3D and the best recall on Tanks \& Temples.}
\label{fig1}
\end{figure}

One notable method, APD-MVS \cite{Wang2023}, introduces adaptive patch deformation. This method classifies pixels into reliable and unreliable based on matching ambiguity. For each unreliable pixel, it searches for a number of reliable pixels in the surrounding area. The RANSAC algorithm is then used to estimate the best-fitting plane from these reliable pixels, selecting the most fitting ones as anchors to assist in the matching of unreliable pixel. Compared to previous methods, this approach is more flexible and significantly enhances the robustness of the plane model.

Although these methods significantly improve reconstruction in low-textured areas, they still face issues with increased scene complexity, as shown in Fig. 2 and Fig. 3. One common issue is the plane model crossing object boundaries, causing depth confusion between foreground objects and the background. Another is the potential for errors in plane model construction due to limited perception range. For example, APD-MVS considers only the nearest reliable pixels when constructing planes, sometimes selecting locally optimal pixels, which leads to noticeable deviations between the final plane model and the ground truth.

To address these issues, we drew inspiration from learning-based MVS methods \cite{Zhang2023, Li2024}, which utilize RGB images for adaptive sampling. We posit that fully leveraging image information, particularly edge information, is crucial since areas delineated by edges often approximate planar shapes. Based on this premise, we integrated dual-level precision edge information into the adaptive patch deformation. Dual-level precision edges are derived from two edge detection 
approaches: fine edges, which are precise but incomplete, and coarse edges, which capture more actual object boundaries but with less accuracy. Specifically, we use the Canny operator for fine edges and a segmentation scheme from TSAR-MVS \cite{Yuan2024a} for coarse edges. Fine edges constrain point selection during plane construction, while coarse edges expand the perception range for selecting anchors. Thereby providing more effective support for the matching of unreliable pixels. Furthermore, since reliable pixels serve as the basis for selecting anchors and are still processed using conventional PM in APD-MVS, we utilize fine edge information to improve the hypotheses sampling strategy, optimizing the hypotheses for reliable pixels.
\begin{figure}[t]
\centering
\includegraphics[width=1\columnwidth]{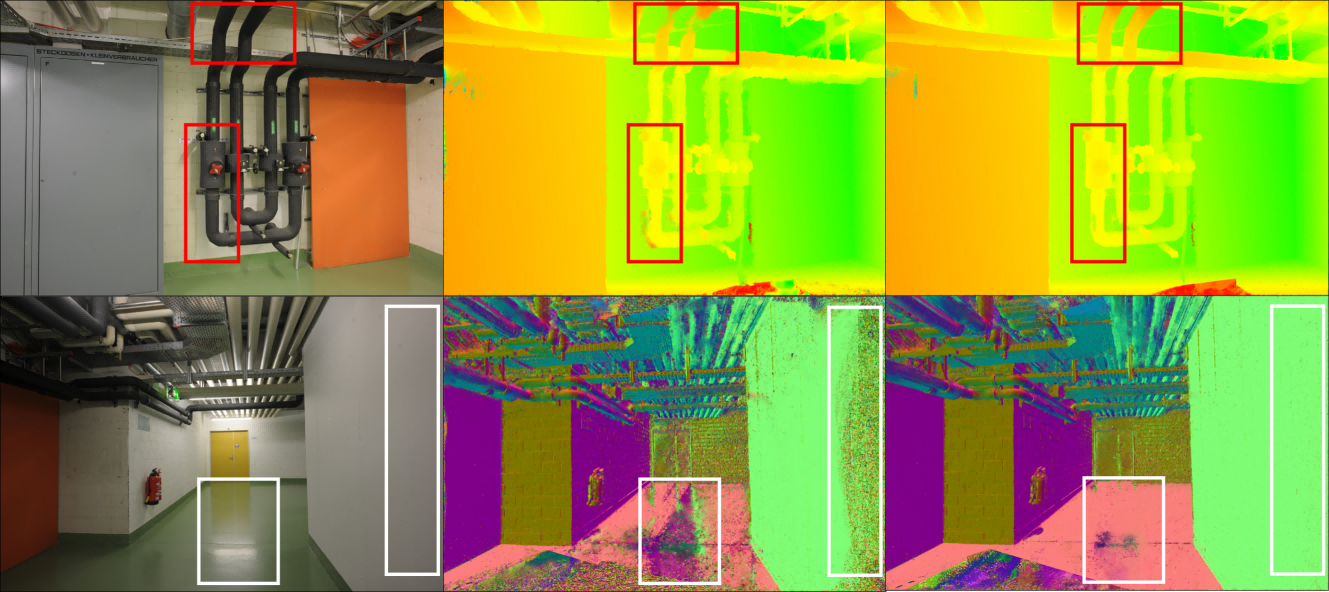} % Reduce the figure size so that it is slightly narrower than the column.
\caption{Top: depth maps for scenes crossing object boundaries. Bottom: normal maps for limited perception range. Comparison of APD-MVS (middle) and our method (right).}
\label{fig2}
\end{figure}

The aforementioned strategy has shown marked performance but is ineffective in stochastic textured areas, such as lawns, which often contain numerous erroneous edges. Therefore, we further investigated the matching cost calculation. The deformable patch, consisting of an unreliable pixel's patch and the anchors' patches, is used to evaluate the plane hypothesis of the the unreliable pixel. Its matching cost is calculated as the weighted sum of the matching costs of both the unreliable pixel's patch and the anchors' patches. However, the conventional fixed-size patch matching method, specifically the matching of the unreliable pixel’s patch, remains part of this process, affecting stability, especially in stochastic textured areas. To address this, we propose adjusting patch sizes with anchors to effectively identify unreliable pixels and applying edge information constraints to prevent crossing object boundaries. This enables more robust matching cost calculations.

\begin{figure}[t]
\centering
\includegraphics[width=1\columnwidth]{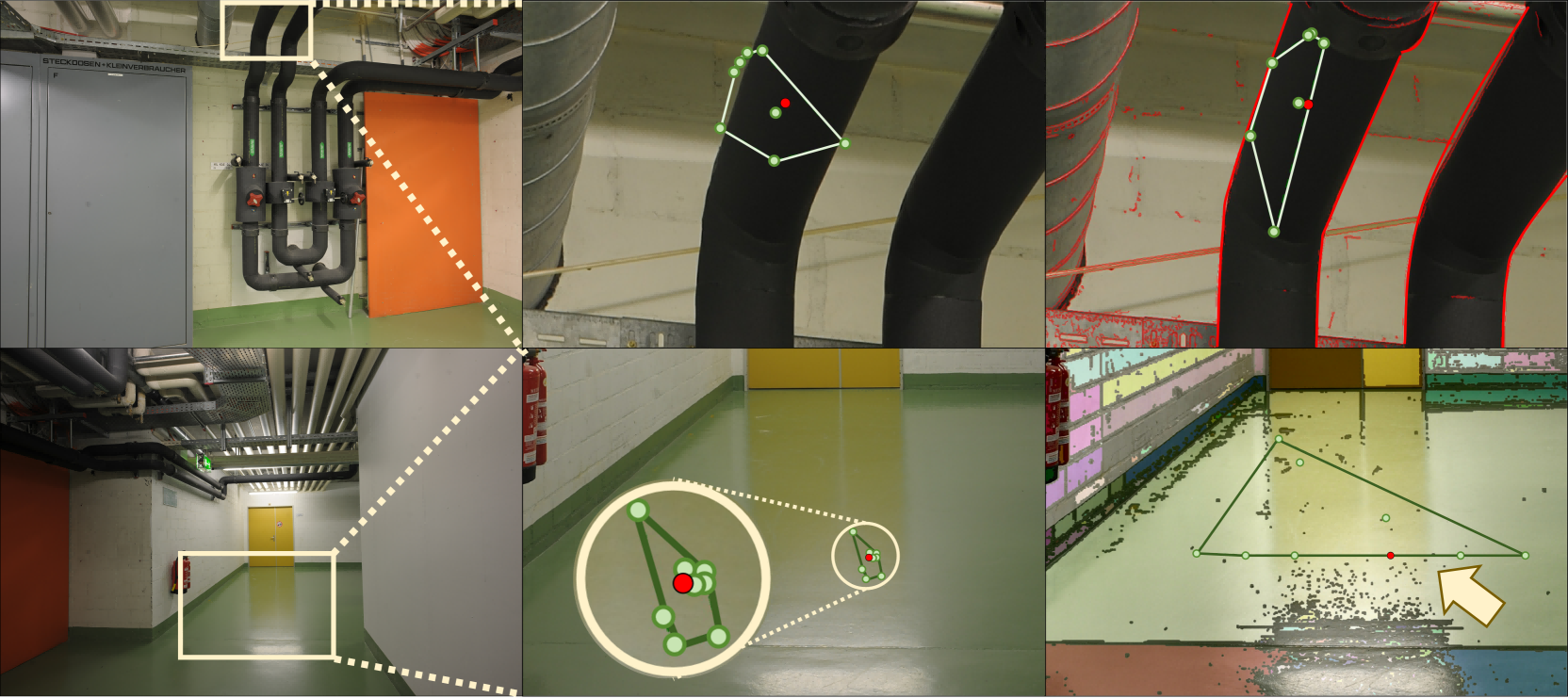} % Reduce the figure size so that it is slightly narrower than the column.
\caption{Comparison of plane construction between APD-MVS (middle) and our method (right), with visualizations of fine edges (top right) and coarse edges (bottom right).}
\label{fig3}
\end{figure}

We integrated the aforementioned concepts into the Dual-Level Precision Edges Guided Multi-View Stereo with Accurate Planarization (DPE-MVS). In summary, our contributions can summarized as follows:

\begin{itemize}
\item We propose a dual-level precision edge-guided planar model construction strategy, providing more effective support for the matching of unreliable pixels.
\item We propose a sampling strategy guided by fine edges, which can enhance conventional PM for reliable pixels.
\item We introduce an adaptive patch size adjustment approach that enables more robust matching cost calculation for unreliable pixels.
\item Extensive experiments validate the effectiveness of our proposed method, demonstrating state-of-the-art performance on the ETH3D and Tanks \& Temples benchmarks.
\end{itemize}

\begin{figure*}[t]
\centering
\includegraphics[width=1\textwidth]{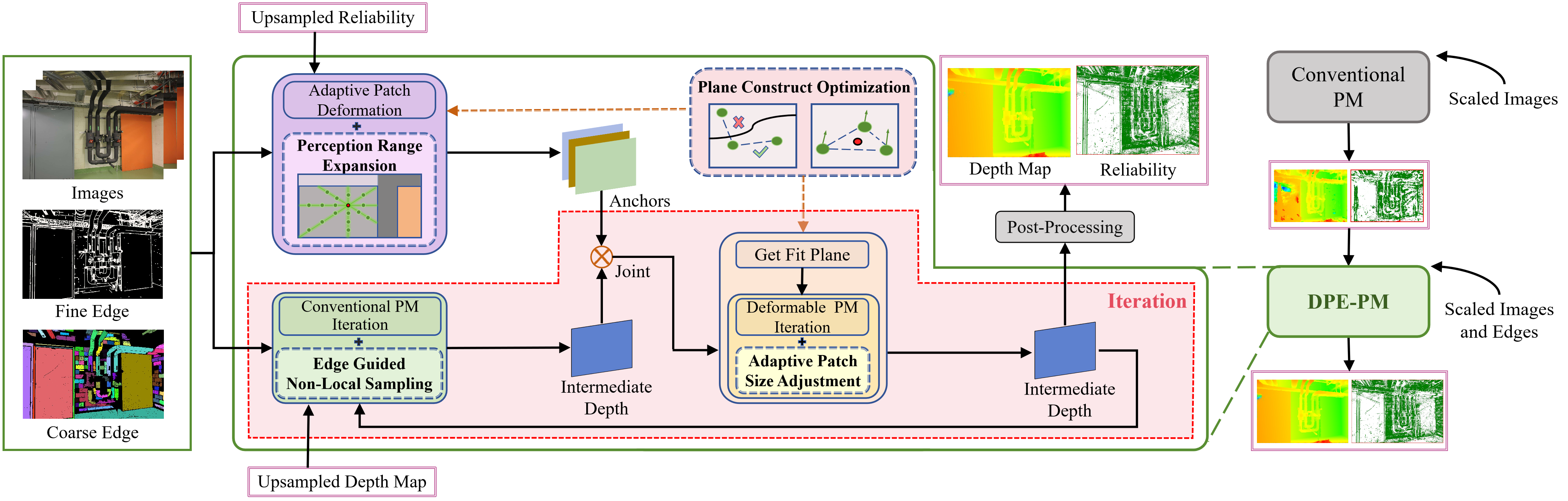} % Reduce the figure size so that it is slightly narrower than the column.
\caption{Overview. DPE-MVS adopt a pyramid structure, with the two coarsest scales displayed on the right side of the figure, and the middle illustrating the details of our proposed DPE-PM. Iterations at finer scales use DPE-PM to update the depth map.}
\label{fig4}
\end{figure*}

\section{Related Work}
\textbf{Traditional Methods} Traditional MVS methods can be roughly categorized into four types: voxel-based methods \cite{Vogiatzis2007}, surface iterative optimization methods \cite{Cremers2010}, patch-based methods \cite{Furukawa2009}, and depth map-based methods \cite{Bleyer2011}. Among these, depth map-based methods have become the most popular choice in recent years due to their simplicity, flexibility, and robust performance. Many outstanding works within this category are PM-based methods. Recently, methods such as ACMM \cite{Xu2019}, ACMP \cite{Xu2020}, and ACMMP \cite{Xu2022} have introduced pyramid structures, geometric consistency, and triangular plane priors into MVS. Subsequently, HPM-MVS \cite{Ren2023} proposed non-local sampling to escape local optima and used a KNN-based approach to optimize plane prior model construction. APD-MVS \cite{Wang2023} introduced adaptive patch deformation and an NCC-based matching metric to determine the reliability of pixel depth values. Methods like TAPA-MVS \cite{Romanoni2019} and PCF-MVS \cite{kuhn2019plane} incorporated superpixel segmentation and the RANSAC algorithm, while TSAR-MVS \cite{Yuan2024a} further combined the Roberts operator with Hough line detection to segment large low-textured areas, though these methods tend to over-segment. SD-MVS \cite{Yuan2024} used SAM for semantic segmentation to achieve adaptive sampling. However, SAM’s inference speed is slow and it may produce errors with unseen scenes. Additionally, SAM struggles to distinguish different surfaces of the same object, making it unsuitable for constructing plane models.

\medskip
\noindent\textbf{Learning-based Methods} MVSNet \cite{Zhang2023} pioneered the use of deep learning for depth map-based MVS methods. CasMVSNet \cite{Gu2020} introduced a cascade structure, accelerating the evolution of learning-based MVS methods. These methods, benefiting from convolution operations, have significantly larger receptive fields compared to traditional methods. Works like AA-RMVSNet \cite{Wei2021} and TransMVSNet \cite{Ding2022} further expanded the receptive field. Additionally, improvements in depth sampling have been made by PatchMatchNet \cite{Wang2021} and DS-PMNet \cite{Li2024}, which proposed adaptive hypothesis propagation, and N2MVSNet \cite{Zhang2023}, which introduced adaptive non-local sampling and RGB-guided depth refinement. Learning-based methods have stronger feature perception and often outperform traditional methods with sufficient data. However, creating high-quality datasets remains challenging, limiting practical use.

\section{Method}
Given a set of images $\{I_i\}^N_{i=1}$ and the corresponding camera parameters $\{\mathbf{P}_i\}^N_{i=1}$, our task is to estimate depth maps for each image. This section provides a brief overview of the key points of APD-MVS, followed by a detailed explanation of our method.

\subsection{Review of APD-MVS}
APD-MVS classifies pixels as reliable or unreliable based on matching ambiguity and introduces deformable PM. Reliable pixels are processed using conventional PM, while unreliable pixels are handled using deformable PM.

Conventional PM consists of four basic steps: random initialization, hypothesis propagation, multi-view matching cost evaluation, and refinement. First, each pixel is randomly initialized with a plane hypothesis. Second, hypotheses are sampled from neighboring pixels within a fixed range. Third, matching costs from multiple views are integrated to select the best hypothesis. Fourth, new hypotheses are generated through perturbation and random 
generation to diversify the solution space, and the best one is selected. The last three steps iterate multiple times.

Deformable PM differs from conventional PM in propagation and matching cost calculation. For each unreliable pixel, anchors are identified through preprocessing. In  propagation, sampled hypotheses include anchor hypotheses and plane hypothesis generated using RANSAC on the anchors. 
% RANSAC (Random Sample Consensus) iteratively selects random subsets of three data points to fit a plane, identifying the subset with the most inliers as the best fit. 
In matching cost calculation, the deformable patch is constructed by combining the unreliable pixel's patch with the anchors' patches for matching, the formula as follows:
\begin{equation}\label{eq1}
m_D\mkern-1mu(\mathbf{p}, \boldsymbol{\theta}_p,\mathbf{S})=\lambda m\mkern-1mu(\mathbf{p}, \boldsymbol{\theta}_p,\mathbf{B}_p)+\frac{1-\lambda}{|\mathbf{S}|}\sum_{\mathbf{s} \in \mathbf{S}}m\mkern-1mu(\mathbf{s},\boldsymbol{\theta}_p,\mathbf{B}_s),\mkern-2mu
\end{equation}
where $\lambda$ is a weight value, $\mathbf{p}$ represents the unreliable pixel, S denotes the set of anchors. $\boldsymbol{\theta}_p$ is the plane hypothesis for pixel $\mathbf{p}$, and $\mathbf{B}$ represent the fixed-size patch. The function $m$ represents the conventional matching cost, while $m_D$ denotes the matching cost of the deformable patch.

\subsection{Overview of Our Method}
Our method adopts the APD-MVS framework, and an overview is illustrated in Fig. 4. 
Each image is sequentially taken as the reference image $I_{ref}$, with the other images as source images $I_{src}$ to guide the reference image's depth map recovery. We construct an $L$ layer pyramid structure through scale downsampling, with the $L$-th layer as the coarsest scale and the 1st layer as the original image. The initial depth map at the coarsest scale layer is obtained using conventional PM, and perform post-processing to determine the reliability of each pixel. At a finer scale layer $l$, fine and coarse edges are extracted from $I_{ref}$ at the corresponding scale. The depth map and reliability from layer $l+1$ are upsampled. These inputs are then used to update the depth map and reliability for this layer using DPE-PM.

In DPE-PM, the first stage is to obtain anchors for each unreliable pixel. In adaptive patch deformation, we propose Perception Range Expansion to search for a wide range of relevant reliable pixels and use RANSAC to filter out the anchors. The second stage is to iteratively update the depth map. In each iteration, the hypotheses for reliable pixels are first updated using conventional PM with our Edge Guided Non-Local Sampling. Subsequently, the hypotheses for unreliable pixels are updated: new plane hypothesis are generated using RANSAC based on the anchors, followed by deformable PM with our Adaptive Patch Size Adjustment. In the previously mentioned RANSAC applications, our Plane Construction Optimization was consistently utilized to obtain accurate plane models. Similarly, post-processing is used to determine pixel reliability after obtaining the depth map. The DPE-PM process is repeated at finer scales until the depth map for the first layer is 
obtained. Finally, the depth maps are fused to generate a point cloud.

In the following sections, we will provide the details of our method. First, edge extraction will be introduced, followed by the improvements related to reliable pixels, and finally, the improvements related to unreliable pixels.

\subsection{Extracting Edge Cues}
Obtaining edge information is essential for our method. We first use the Canny edge detector to extract fine edges, setting the upper and lower thresholds to the median of the image grayscale multiplied by $(1 \pm \sigma)$. Then, coarse edges are extracted using the Roberts operator and Hough line detection, similar to TSAR-MVS. Fine edges help accurately locate the boundaries of foreground objects but are often not closed, making it difficult to determine whether a pixel is in a low-textured region. Coarse edges can segment the image, with larger regions indicating low-textured areas, but have a higher false detection rate and less precise boundary localization. The complementary information from fine and coarse edges is crucial for achieving our research objectives.

\subsection{Edge Guided Non-Local Sampling} 

Reliable pixels are the foundation for constructing the plane model, and the accuracy of their hypotheses is crucial. 
% Insights from \cite{Ren2023} suggest that ignoring local sampling during the PM propagation process can expand the search range of the solution space, while \cite{Zhou2021} emphasizes the importance of local samples for detail retention. % 
Propagation samples hypotheses from neighboring pixels to build the solution space. According to \cite{Ren2023, Zhou2021}, repetitive hypotheses often occur within the local range in low-textured areas, whereas local sampling preserves fine details in small objects.
To expand the solution space while retaining details, we propose two new sampling schemes: progressive non-local sampling and edge-guided extended sampling, as shown in Fig. 5. For fine edge pixels, we apply only progressive non-local sampling. For non-fine edge pixels, both sampling schemes are applied, and the resulting samples are compared to retain the superior ones.
% This strategy includes two sub-strategies. First, we apply a progressive non-local sampling strategy. Second, it is determined whether the pixel currently being processed is a fine edge pixel. If it is not, an additional fine edge-guided extended sampling strategy is applied.

\begin{figure}[t]
\centering
\includegraphics[width=1\columnwidth]{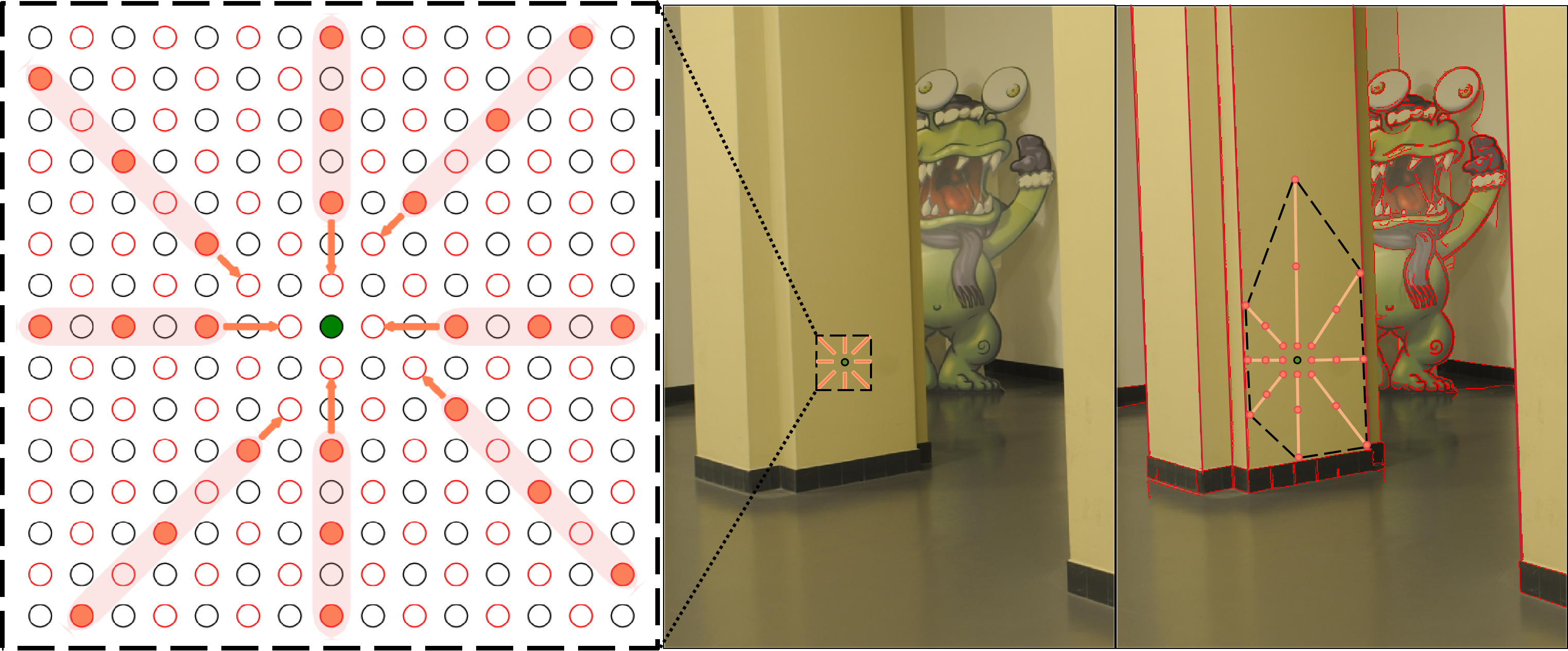} % Reduce the figure size so that it is slightly narrower than the column.
\caption{Edge Guided Non-Local Sampling: This process includes two sampling schemes: progressive non-local sampling (left) and edge-guided extended sampling (right).}
\label{fig5}
\end{figure}

Progressive non-local sampling excludes sampling points within a radius $\xi$ during the PM iteration process. The radius $\xi$ gradually decreases as the iterations progress, following the formula $\xi = max(1, 5 - 2 \times t_{iter})$. We utilize a red-black checkerboard pattern for pixel division \cite{Xu2019}. Each of the eight sampling areas adopts a strip format, where the radius $\xi$ offsets the starting position of the strip. Each area contains 11 samples with a step size of 2. The sample with the minimum multi-view matching cost is selected from each area, yielding 
the optimal samples $\{\boldsymbol{\theta}_i^{pn}\}_{i=1}^8$.

Edge-guided extended sampling follows the same fundamental scheme as progressive non-local sampling, but differs in the number of samples and the sampling step size. In this scheme, both the number of samples $k$ and the step size $s$ are adaptively adjusted based on the distance $D_{fe}$ to the nearest fine edge in the corresponding strip direction. A threshold, defined as $\Lambda_{fe} = \frac{imagewidth}{30 \times 4^l}$, is used to prevent excessively large sampling distances. The specific adjustments are calculated as follows:
\begin{equation}
    D_{fe}^{\prime} = min\left(D_{fe}, \Lambda_{fe}\right),
    k = \left\lfloor\frac{D_{fe}^{\prime}}{2}\right\rfloor,
    s = \left\lfloor\frac{D_{fe}^{\prime}}{k}\right\rfloor,
\end{equation}
 subject to $11 \leq k \leq 22$. Similarly, optimal samples $\{\boldsymbol{\theta}_i^{eg}\}_{i=1}^8$ are obtained. These samples are then compared with
$\{\boldsymbol{\theta}_i^{pn}\}_{i=1}^8$ by recalculating the matching costs on the patch of the pixel being processed, with the better sample for each area direction being selected.

\subsection{Accurate Plane Model Construction}
For unreliable pixels, the anchors filtered out by planar RANSAC in adaptive patch deformation significantly impact depth estimation.

\medskip
\noindent\textbf{Perception Range Expansion} To address the issue of limited perception range, we segment the image into distinct regions using coarse edges and extend the search for reliable pixels in low-textured regions. A region is considered low-textured if its pixel count exceeds $\frac{imagearea}{256 \times 4^l}$. Let $\boldsymbol{\mathcal{E}}$ denote the set of coarse edge pixels in a specific direction relative to the pixel $\mathbf{p}$, the boundary in that direction is defined as:
\begin{equation}
    \mathcal{B}(\mathbf{p}) = \operatorname*{arg\,max}_{\mathbf{q} \in \boldsymbol{\mathcal{E}}} ||\mathbf{q}-\mathbf{p}|| \quad s.t. \quad\!\! \mathbb{C}(\mathbf{q}) = 1,    
\end{equation} where $\mathbb{C}(\mathbf{q}) = 1$ indicates that $\mathbf{q}$ is connected to the region where 
$\mathbf{p}$ is located. This approach determines boundaries in the eight-connected directions for each pixel, effectively filtering out redundant coarse edge pixels within the region.

Subsequently, the eight directions are grouped into four pairs of opposites. Each pair has a search limit of $2\eta$ reliable pixels, distributed between directions based on boundary distances. For example, in the up-down direction pair, the number of pixels allocated for searching is determined using the following formula:
\begin{equation}
    n_{u} = \left\lfloor\frac{2\eta \cdot D_{ce}^{u}}{D_{ce}^{u} + D_{ce}^{d}}\right\rfloor, n_{d} = 2\eta - n_{u},
\end{equation}
where $D_{ce}^{u}$ and $D_{ce}^{d}$ represent the boundary distances in the upward and downward directions, respectively, with $n_u$ is constrained to $1 \leq n_u \leq 2\eta - 1$. For each direction, starting from the unreliable pixel, we locate a set of equally spaced pixels $\{\mathbf{s}_i\}_{i=1}^n$ along the line to the boundary. Let $\mathcal{N}(\mathbf{s}_i)$ denote the nearest reliable pixel to $\mathbf{s}_i$. The set $\{\mathcal{N}(\mathbf{s}_i)\}_{i=1}^n$ serves as the result of 
extended search in that direction.

We retain the reliable pixels search scheme from APD-MVS, which involves partitioning the search space centered on the unreliable pixel into $\phi$ equal-angle sectors and locating the nearest reliable pixels within each sector. This scheme only searches the nearest reliable pixels and is applied to each unreliable pixel, while the extended search supplements it in low-textured areas. 

\medskip
\noindent\textbf{Plane Construction Optimization} RANSAC aims to find the best-fitting plane from 3D points. Using an unreliable pixel in a low-textured area as an example, obtain the set of 3D points $\boldsymbol{\mathcal{X}} = \{\mathbf{X}_i\}^{\phi+8\eta}_{i=1}$ for the searched reliable pixels. Three random points from $\boldsymbol{\mathcal{X}}$ are iteratively selected to construct a plane $\boldsymbol{\pi}$, identifying the best fit as follows:
\begin{equation}
     \boldsymbol{\pi}^* = \operatorname*{arg\,min}_{\boldsymbol{\pi}}  \sum^{\phi+8\eta}_{i=1} \mathbb{I}[\delta(\mathbf{X}_i, \boldsymbol{\pi}) < \epsilon].
\end{equation}
Here, $\delta$ represents the residual, $\epsilon$ is the outlier filtering threshold, and $\mathbb{I}$ is an indicator function that equals 1 if the condition is met and 0 otherwise. If there are multiple $\boldsymbol{\pi}^*$, select the one that minimizes $\delta$ with the 3D point of the unreliable pixel. Let $\boldsymbol{\pi}$ be defined by $m_1x + m_2y + m_3z + b = 0$. Based on the projection relationship between $\mathbf{X}_i$ and its pixel coordinate $(u_i, v_i)$, the fitting depth $d_i$ is obtained as: 

\begin{equation}
    d_i = \frac{-b \cdot f_x \cdot f_y}{m_1f_y(u_i-c_x)+m_2f_x(v_i-c_y)+m_3f_xf_y},
\end{equation}
where $c_x$, $c_y$, $f_x$ and $f_y$ are intrinsic camera parameters.
For $\mathbf{X}_i$, we calculated $\delta$ as $|d_i - d_i^\theta|$, where $d_i^\theta$ is the depth hypothesis. Comparatively, conventional planar RANSAC calculates $\delta$ by considering only the point-to-plane distance, without utilizing the projection relationship.

To mitigate the issue of crossing object boundaries, we imposed constraints on RANSAC's random point selection. Let $\boldsymbol{\mathcal{F}}$ denote the set of all fine edge pixels, $\{\mathbf{q}_i,\mathbf{n}_i^\theta\}_{i=1}^3$ represent the pixels and their normal hypotheses corresponding to the selected three points. We present two conditions:
\begin{equation}
\forall i, j \in \{1, 2, 3\},i<j \quad
\begin{aligned}
& C_1: \quad \overline{\mathbf{q}_i \mathbf{q}_j} \cap \boldsymbol{\mathcal{F}} = \emptyset; \\
& C_2: \quad \mathbf{n}_i^\theta \cdot \mathbf{n}_j^\theta > \tau,
\end{aligned}
\end{equation}
where $\overline{\mathbf{q}_i \mathbf{q}_j}$ represents the set of pixels along the line segment between $\mathbf{q}_i$ and $\mathbf{q}_j$, and $\tau$ is a threshold. Condition $C_1$ is a mandatory requirement, while $C_2$ is a priority condition that should be met when possible.

After RANSAC, select up to $|\mathbf{S}|$ pixels corresponding to the 3D points $\{\mathbf{X}_i | \delta(\mathbf{X}_i, \boldsymbol{\pi}^*) < \epsilon\}$ that have smallest residuals to serve as anchors. The optimized RANSAC is also used to generate a new plane hypothesis with these anchors, a necessary step for the propagation in deformable PM.

\begin{figure}[t]
\centering
\includegraphics[width=1\columnwidth]{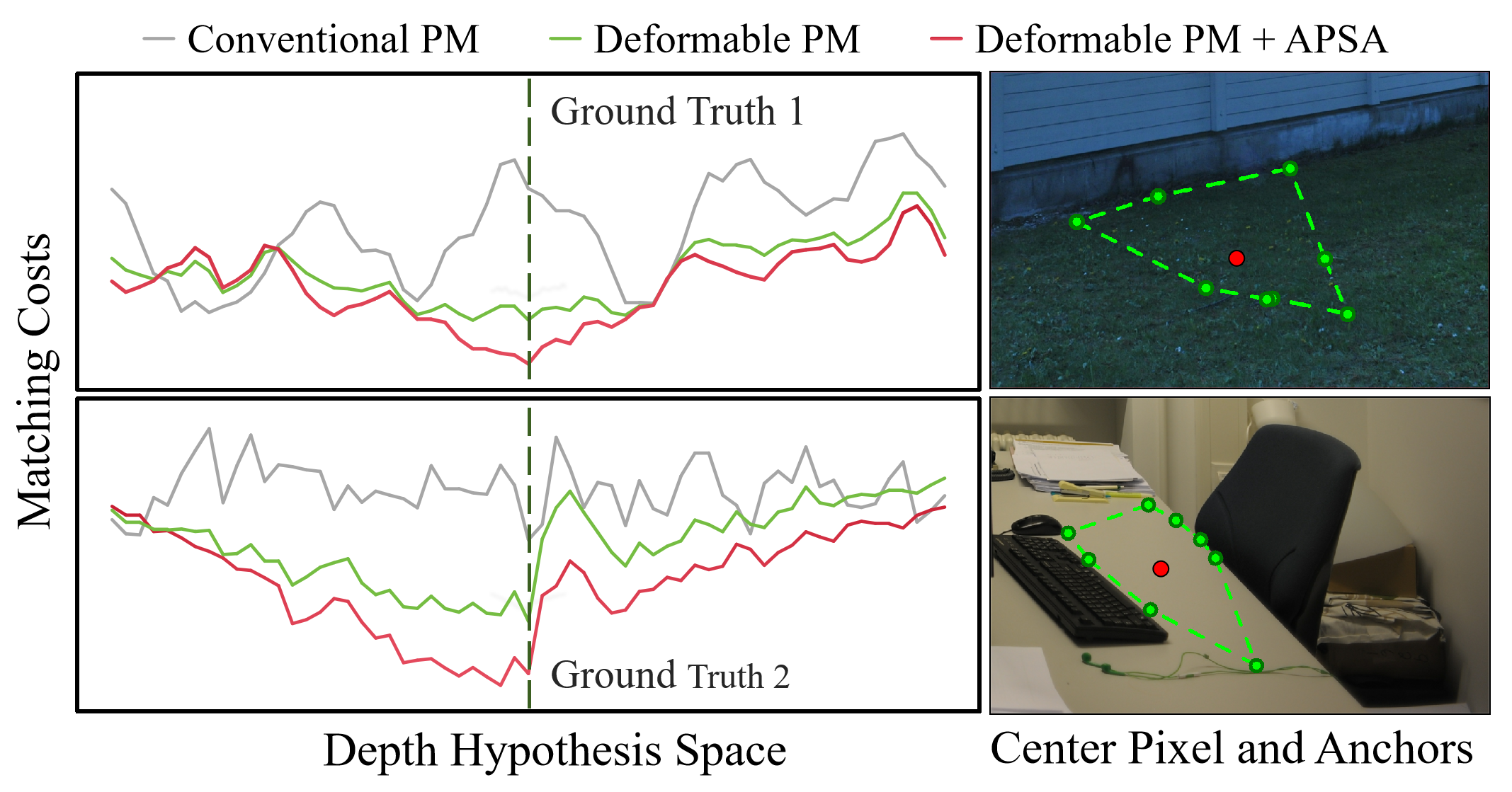} % Reduce the figure size so that it is slightly narrower than the column.
\caption{The right side presents examples of a stochastic textured area (top) and a low-textured area (bottom). The left side shows the corresponding cost profiles, highlighting the matching costs around the ground truth for various methods, including conventional PM, deformable PM, and deformable PM with our Adaptive Patch Size Adjustment.}
\label{fig6}
\end{figure}

\begin{table*}[!ht]
\centering
% \resizebox{.95\textwidth}{!}{
\fontsize{9pt}{10pt}\selectfont
\begin{tabular}{c|c c c|c c c|c c c|c c c}
\hline 
% \multicolumn{2}{c|}
{\multirow{3}*{Method}}

 & \multicolumn{6}{|c|}{Train} & \multicolumn{6}{|c}{Test} \\
\cline{2-13}

% \multicolumn{2}{c|}{} 
& 
\multicolumn{3}{|c|}{2cm} & \multicolumn{3}{|c|}{10cm} & \multicolumn{3}{|c|}{2cm} & \multicolumn{3}{|c}{10cm} \\
\cline{2-13}

% \multicolumn{2}{c|}{} 
& Acc. & Comp. & F$_1$ & Acc. & Comp. & F$_1$ & Acc. & Comp. & F$_1$ & Acc. & Comp. & F$_1$ \\ 
\hline 

% \multirow{4}*{\rotatebox[origin=c]{90}{Learning}}

PatchMatchNet
% $_{\scriptsize{C\hspace{-0.1em}V\hspace{-0.3em}P\hspace{-0.2em}R2021}}$ 
& 64.81 & 65.43 & 64.21 & 89.98 & 83.28 & 85.70 & 69.71 & 77.46 & 73.12 & 91.98 & 92.05 & 91.91 \\
IterMVS
% $_{\scriptsize{C\hspace{-0.1em}V\hspace{-0.3em}P\hspace{-0.2em}R2022}}$ 
& 79.79	& 66.08	& 71.69	& 96.35	& 82.62	& 88.60	& 84.73	& 76.49	& 80.06 & 96.92 & 88.34 & 92.29 \\
EPNet
% $_{A\!A\!A\!I2023}$ 
& 79.36 & 79.28 & 79.08 & 94.33 & 93.69 & 93.92 & 80.37 & \underline{87.84} & 83.72 & 93.72	& 96.82 & 95.20 \\
GoMVS
% $_{\scriptsize{C\hspace{-0.1em}V\hspace{-0.3em}P\hspace{-0.2em}R2024}}$ 
& 81.22 & 77.65 & 79.16 & 97.11 & 89.62 & 93.08 & 86.85 & 85.50 & 85.91 & 97.23 & 95.02 & 96.02 \\
\hline 

% \multirow{7}*{\rotatebox[origin=c]{90}{Traditional}}
% & COLMAP
% % $_{\scriptsize{C\hspace{-0.1em}V\hspace{-0.3em}P\hspace{-0.2em}R2016}}$ 
% & \textbf{91.85} & 55.13 & 67.66 & \textbf{98.75} & 79.47 & 87.61 & \underline{91.97} & 62.98 & 73.01 & \textbf{98.25} & 84.54 & 90.40 \\
ACMM
% $_{\scriptsize{C\hspace{-0.1em}V\hspace{-0.3em}P\hspace{-0.2em}R2019}}$ 
& \textbf{90.67} & 70.42 & 78.86 & \underline{98.12} & 86.40 & 91.70 & 90.65 & 74.34 & 80.78 & \underline{98.05} & 88.77 & 92.96 \\
ACMMP
% $_{T\!P\!A\!M\!I2022}$ 
& 90.63 & 77.61 & 83.42 & 97.99 & 93.32 & 95.54 & \underline{91.91} & 81.49 & 85.89 & \underline{98.05} & 94.67 & 96.27 \\
TSAR-MVS + MP.
& 89.67 & 84.39 & 86.88 & \textbf{98.15} & 96.50 & 97.31 & 88.14 & 88.11 & 88.02 & 97.42 & 97.44 & \underline{97.42} \\
APD-MVS
% $_{\scriptsize{C\hspace{-0.1em}V\hspace{-0.3em}P\hspace{-0.2em}R2023}}$ 
& 89.14 & \underline{84.83} & 86.84 & 97.47 & 96.79 & 97.12 & 89.54 & 85.93 & 87.44 & 97.00 & 96.95 & 96.95 \\
HPM-MVS
% $_{\scriptsize{I\hspace{-0.1em}C\hspace{-0.1em}C\hspace{-0.1em}V\hspace{-0.1em}2023}}$ 
& \underline{90.66} & 79.50 & 84.58 & 97.97 & 95.59 & 96.22 & \textbf{92.13} & 83.25 & 87.11 & \textbf{98.11} & 95.41 & 96.69 \\
SD-MVS
% $_{\scriptsize{A\hspace{-0.1em}A\hspace{-0.1em}A\hspace{-0.1em}I2024}}$
& 89.91 & 84.31 & \underline{86.96} & 97.84 & \underline{96.87} & \underline{97.35} & 88.96 & 87.49 & \underline{88.06} & 97.37 & \underline{97.51} & 97.41 \\
\hline
DPE-MVS (ours) & 89.81 & \textbf{87.60} & \textbf{88.63} & 97.99 & \textbf{97.69} & \textbf{97.83} & 90.53 & \textbf{88.77} & \textbf{89.48} & 97.64 & \textbf{98.11} & \textbf{97.86} \\
\hline

\end{tabular}
% }
% }
\caption{Quantitative results on ETH3D benchmark. Our method achieves the best completeness and F$_1$-score.}
\label{table1}
\end{table*}

Moreover, we observed that stochastic textured areas, lacking explicit repeating patterns \cite{efros1999texture}, often lead to numerous fine edges that do not effectively constrain the plane model. To address this, we propose assessing texture complexity per pixel. Let 
$N_{fe}$ and $N_{ce}$ denote the number of fine and coarse edge pixels, respectively,  within an $M \times M$ area centered on pixel $\mathbf{p}$.
 The probability that $\mathbf{p}$ is in a stochastic textured area can be calculated as:
\begin{equation}
    \label{eq5}
    \Phi_p = \frac{\alpha \cdot N_{fe} + (1-\alpha) \cdot N_{ce}}{M \times M},
\end{equation}
\begin{equation}
    \label{eq6}
    \mathcal{P}(\mathbf{p}|Z)=\frac{1}{1+exp\left[-\beta_1 \cdot \left(\Phi_p - \beta_2\right)\right]},
\end{equation}
where $\Phi_p$ represent edge densities, and $\alpha$, $\beta_1$, $\beta_2$ are empirically set. $Z$ indicates that the pixel is in a stochastic textured area, and $M$ is the same as the fixed patch size. We generate a random number $r$ between 0 and 1. If $r < \mathcal{P}(\mathbf{p}|Z)$, then $\mathbf{p}$ is considered to be in a stochastic textured area, rendering condition $C_1$ inapplicable.

\subsection{Adaptive Patch Size Adjustment}

When computing matching cost, the deformable patch employs fixed-size patches of the center unreliable pixel and its anchors, as detailed in Eq. 1. The cost profiles for different matching methods in stochastic and low-textured areas are shown in Fig. 6. The gray cost profiles indicate the instability of the center patch during matching, suggesting the need for further improvement. Enlarging the center patch size may help mitigate this issue \cite{Xu2020a, Sun2022}, but could lead to detail loss and challenges in selecting the size.

\begin{algorithm2e}[t]
\vfill
%\SetAlgoLined		% 增添end行
\DontPrintSemicolon
\SetKwInOut{Input}{\textbf{Input}}		% Set the Input
\SetKwInOut{Output}{\textbf{Output}}	% set the Output

\Input{the three anchors that define the plane for unreliable pixel $\mathbf{p}$, fine and coarse edges}
\Output{appropriate center patch radius $\gamma$ for $\mathbf{p}$}
 $\gamma \gets \lfloor\sqrt{triangle area} / 2\rfloor$; \\
\tcp{the three anchors}
$\boldsymbol{D} \gets \{\omega \times D_{an}\}$; \\
\If{$\mathbf{p}$ is not in a stochastic textured area} {
    \tcp{eight fine edges}
    $\boldsymbol{D} \gets \boldsymbol{D} \cup \{D_{fe}\}$; \\
    \If{$\mathbf{p}$ is in a low-textured area} {
    \tcp{eight coarse edges}
        $\boldsymbol{D} \gets \boldsymbol{D} \cup \{D_{ce}\}$; \\
    }
}
$\gamma \gets min\{ \gamma \cup \boldsymbol{D}\}$; \\
    \tcp{fixed patch size}
    \If{$fixed radius > \gamma$} { $\gamma \gets fixed radius$; }
\ElseIf{$\mathbf{p}$ is in a stochastic textured area} {
    $\gamma \gets 0$;
}
\textbf{return} $\gamma$;
\caption{Adaptive Patch Size Adjustment}
\end{algorithm2e}

% \begin{algorithm2e}[ht]
% \vfill
% %\SetAlgoLined		% 增添end行
% \DontPrintSemicolon
% \SetKwInOut{Input}{\textbf{Input}}		% Set the Input
% \SetKwInOut{Output}{\textbf{Output}}	% set the Output

% \Input{three anchors that define the plane for pixel $\mathbf{p}$, fine and coarse edges}
% \Output{appropriate center patch radius $\gamma$ for $\mathbf{p}$}
%  $\gamma \gets \lfloor\sqrt{triangle area} / 2\rfloor$; \\
% $\mathcal{D}_{an} \gets min\{D_{an}\}$; \\
% \If{$\omega \times \mathcal{D}_{an} < \gamma$} {
%     $\gamma \gets \mathcal{D}_{an}$;
% }
% \If{$\mathbf{p}$ is not in a stochastic textured area} {
%     $\mathcal{D}_{fe} \gets min\{D_{fe}\}$; \\
%     \If{$\mathcal{D}_{fe} < \gamma$} {
%         $\gamma \gets \mathcal{D}_{fe}$;
%     }
%     \If{$\mathbf{p}$ is in a low-textured area} {
%         $\mathcal{D}_{ce} \gets min\{D_{ce}\}$; \\
%         \If{$\mathcal{D}_{ce} < \gamma$} {
%             $\gamma \gets \mathcal{D}_{ce}$;
%         }
%     }
%     % \tcp{fixed patch size}
%     \If{$fixed radius > \gamma$} { $\gamma \gets fixed radius$; }
% }
% \Else {
%     \If{$fixed radius > \gamma$} {
%         $\gamma \gets fixed radius$;
%     } \Else {
%         $\gamma \gets 0$;
%     }
% }
% \textbf{return} $\gamma$;
% \caption{Adaptive Patch Size Adjustment}
% \end{algorithm2e}

We propose a hybrid approach that combines patch enlargement with patch discarding. Roughly speaking, for pixels in stochastic textured
areas, we tend to discard the center patch, while for those in low-textured areas, we tend to enlarge the center patch. The insight behind discarding the center patch is that anchors, selected for their planar characteristics, maintain stable normal estimation, whereas the center patch may introduce instability. The details of this approach are outlined in Algo. 1. It is posited that the triangular area formed by the three anchors, which are selected by optimized RANSAC to represent the points constructing the best-fitting plane, provides informative features for the center pixel. The appropriate patch radius is determined using this triangular area with the surrounding edge information. The comparison between the green and red cost profiles in Fig. 6 illustrates the effectiveness of our approach.  

\section{Experiments}

\begin{figure*}[t]
\centering
\includegraphics[width=1\textwidth]{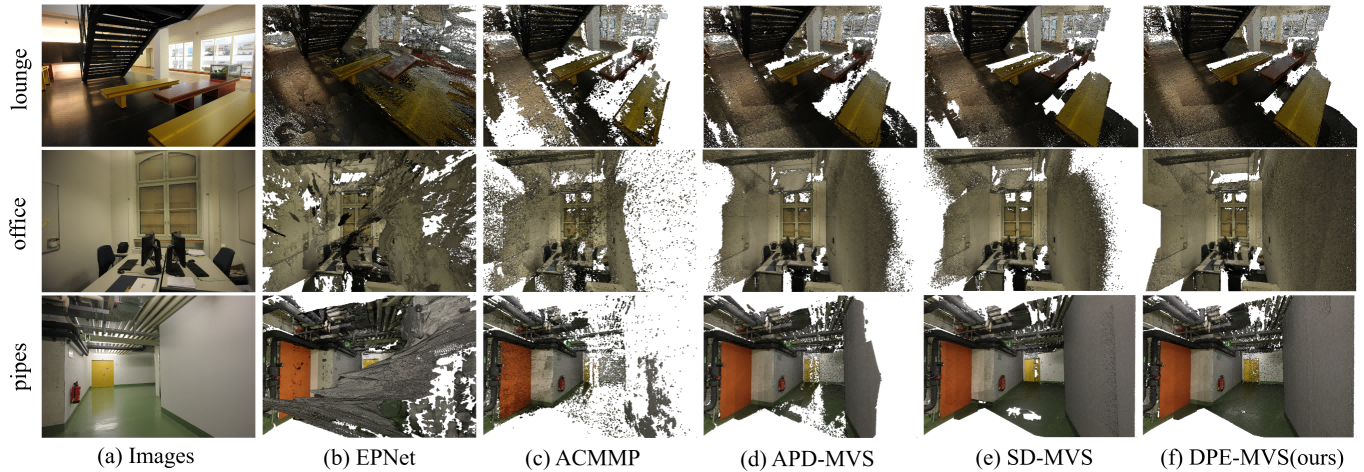} % Reduce the figure size so that it is slightly narrower than the column.
\caption{Qualitative results on ETH3D. Our method exhibits a significant advantage in reconstructing low-textured areas.}
\label{fig7}
\end{figure*}

\subsection{Datasets and Implementation Details} 
We evaluate our method on the ETH3D \cite{Schops2017} and Tanks \& Temples \cite{Knapitsch2017} benchmarks, and conduct ablation experiments on the ETH3D training dataset to verify its effectiveness. We compare our method with SOTA learning-based methods including PatchMatchNet, AA-RMVSNet, IterMVS, TransMVSNet, EPNet, DS-PMNet, GoMVS and traditional MVS methods including ACMM, ACMMP, APD-MVS, HPM-MVS , TSAR-MVS, SD-MVS.
To ensure fairness, we maintained the original APD-MVS parameters and depth map fusion procedures. The proposed parameter settings are: \{$\sigma$, $\eta$, $\tau$, $\alpha$, $\beta_1$, $\beta_2$, $\omega$\} = \{0.67, 4, 0.87, 0.5, 25, 0.35, 2.5\}. 

\subsection{Evaluation on MVS Benchmarks}

The quantitative results of ETH3D are presented in Tab. 1. Our method ranks \textbf{1st} in both completeness and F$_1$-score, with accuracy nearly matching the previous best. Fig. 7 provides a qualitative analysis, confirming that our method significantly enhances the reconstruction of low-textured areas.

On the Tanks \& Temples benchmark, We validated the generalization capability of our method. The quantitative results are shown in Tab. 2. Our method ranks \textbf{1st} in recall among all methods. In comparison with traditional methods, our method's F$_1$-score ranks \textbf{1st} and 2nd in the Intermediate and Advanced datasets, respectively. Compared to learning-based methods, our method is also competitive. It shows lower precision in the Advanced dataset, primarily due to the extensive low-textured areas. In these areas, our method prioritizes recall, with a slight trade-off in precision.

\begin{table}[t]
\centering
\fontsize{9pt}{10pt}\selectfont
% \resizebox{1\columnwidth}{!}{
\setlength{\tabcolsep}{1mm}{
\begin{tabular}{c|c c c|c c c}
\hline
% \multicolumn{2}{c|}
{\multirow{2}*{Method}}

& \multicolumn{3}{|c|}{Intermediate} & \multicolumn{3}{|c}{Advanced} \\
\cline{2-7}
% \multicolumn{2}{c|}{} 
& Pre. & Rec. & F$_1$ & Pre. & Rec. & F$_1$\\ 
\hline 

% \multirow{7}*{\rotatebox[origin=c]{90}{Learning}}

% CasMVSNet
% % $_{\scriptsize{C\hspace{-0.1em}V\hspace{-0.2em}P\hspace{-0.2em}R2020}}$ 
% & 47.62 & 74.01 & 56.84 & 29.68 & 35.24 & 31.12\\
PatchmatchNet
% $_{\scriptsize{C\hspace{-0.1em}V\hspace{-0.3em}P\hspace{-0.2em}R2021}}$ 
& 43.64 & 69.37 & 53.15 & 27.27 & 41.66 & 32.31\\
AA-RMVSNet
% $_{\scriptsize{I\hspace{-0.1em}C\hspace{-0.1em}C\hspace{-0.1em}V2021}}$ 
& 52.68 & 75.69 & 61.51 & \underline{37.46} & 33.01 & 33.53 \\
IterMVS
% $_{\scriptsize{C\hspace{-0.1em}V\hspace{-0.1em}P\hspace{-0.2em}R2022}}$ 
& 47.53 & 74.69 & 56.94 & 28.70 & 44.19 & 34.17 \\
TransMVSNet
% $_{\scriptsize{C\hspace{-0.1em}V\hspace{-0.1em}P\hspace{-0.2em}R2022}}$ 
& 55.14 & 76.73 & 63.52 & 33.84 & 44.29 & 37.00 \\
EPNet
% $_{\scriptsize{A\hspace{-0.1em}A\hspace{-0.1em}A\hspace{-0.1em}I2023}}$ 
& \textbf{57.01} & 72.57 & 63.68 & 34.26 & \underline{50.54} & \underline{40.52} \\
DS-PMNet
% $_{\scriptsize{A\hspace{-0.1em}A\hspace{-0.1em}A\hspace{-0.1em}I2024}}$ 
& \underline{56.02} & 76.76 & \textbf{64.16} & 34.29 & 48.73 & 39.78 \\
\hline 

% \multirow{6}*{\rotatebox[origin=c]{90}{Traditional}}

% COLMAP
% % $_{\scriptsize{C\hspace{-0.1em}V\hspace{-0.1em}P\hspace{-0.1em}R2016}}$ 
% & 43.16 & 44.48 & 42.14 & 33.65 & 23.96 & 27.24 \\
ACMM
% $_{\scriptsize{C\hspace{-0.1em}V\hspace{-0.1em}P\hspace{-0.1em}R2019}}$ 
& 49.19 & 70.85 & 57.27 & 35.63 & 34.90 & 34.02 \\
ACMMP
% $_{\scriptsize{T\hspace{-0.1em}P\hspace{-0.1em}A\hspace{-0.1em}M\hspace{-0.1em}I2022}}$ 
& 53.28 & 68.50 & 59.38 & 33.79 & 44.64 & 37.84 \\
TSAR-MVS + MP.
% $_{\scriptsize{T\hspace{-0.1em}P\hspace{-0.1em}A\hspace{-0.1em}M\hspace{-0.1em}I2022}}$ 
& 53.15 & 75.52 & 62.10 & 33.85 & 48.75 & 38.63 \\
APD-MVS
% $_{\scriptsize{C\hspace{-0.1em}V\hspace{-0.1em}P\hspace{-0.1em}R2023}}$ 
& 55.58 & 75.06 & 63.64 & 33.77 & 49.41 & 39.91 \\
HPM-MVS
% $_{\scriptsize{I\hspace{-0.1em}C\hspace{-0.1em}C\hspace{-0.1em}V2023}}$ 
& 51.58 & 76.92 & 61.39 & \textbf{40.67} & 45.42 & \textbf{40.80} \\
SD-MVS
% $_{\scriptsize{A\hspace{-0.1em}A\hspace{-0.1em}A\hspace{-0.1em}I2024}}$ 
& 53.78 & \underline{77.63} & 63.31 & 35.53 & 47.37 & 40.18 \\
\hline
DPE-MVS (ours) & 54.48 & \textbf{78.16} & \underline{63.98} & 31.37 & \textbf{56.45} & 40.20 \\
\hline 

\end{tabular}
}
\caption{Quantitative results on Tanks \& Temples.}
\label{table2}
\end{table}

\begin{table}[t]
\centering

% \resizebox{1\columnwidth}{!}{
\fontsize{9pt}{10pt}\selectfont
\setlength{\tabcolsep}{1mm}{
\begin{tabular}{c c c c|c c c|c c c}
\hline

\multicolumn{4}{c|}{Settings}  & \multicolumn{3}{|c|}{1cm} & \multicolumn{3}{|c}{5cm} \\
\hline

ES & PE & PO & AA & Acc. & Comp. & F$_{1}$ & Acc. &Comp. & F$_{1}$ \\
\hline

& & & & 80.60 & 71.86 & 75.80 & 95.38 & 93.47 & 94.37 \\

\ding{51} & & & & 79.74 & 74.15 & 76.67 & 94.85 & 93.97 & 94.38\\

& \ding{51} & & & 80.16 & 73.08 & 76.27 & 95.30 & 93.87 & 94.54\\

\ding{51} & \ding{51} & & & 80.19 & 74.71 & 77.13 & 95.43 & 94.36 & 94.81\\

\ding{51} & \ding{51} & \ding{51} & & 80.25 & \textbf{75.79} & \underline{77.79} & 95.45 & 94.71 & 95.05\\

\ding{51} & \ding{51} & & \ding{51} & \underline{80.69} & 74.48 & 77.26 & \underline{95.81} & \underline{94.77} & \underline{95.26}\\

\ding{51} & \ding{51} & \ding{51} & \ding{51} & \textbf{81.01} & \underline{75.74} & \textbf{78.11} & \textbf{96.00} & \textbf{95.10} & \textbf{95.53}\\
\hline 

\end{tabular}
}
\caption{Results of our method with different settings on the ETH3D training dataset; baseline: APD-MVS.}
\label{table3}
\end{table}

\begin{table}[t]
\centering

\fontsize{9pt}{10pt}\selectfont
\begin{tabular}{c|c c c c c}
\hline

Method & 1cm & 2cm & 5cm & 10cm & 20cm\\
\hline

ACMM & 67.58 & 78.86 & 87.68 & 91.70 & 94.41 \\
w/. NESP & \textbf{70.70} & \textbf{81.01} & 88.80 & 92.26 & 94.55 \\
w/. ES (ours) & 69.34 & 80.47 & \textbf{89.17} & \textbf{92.83} & \textbf{95.07} \\
\hline

ACMP & 68.72 & 79.79 & 88.32 & 92.03 & 94.43 \\
w/. NESP & 70.87 & 81.45 & 89.43 & 92.72 & 94.78 \\
w/. ES (ours) & \textbf{72.45} & \textbf{83.01} & \textbf{90.70} & \textbf{93.72} & \textbf{95.53}\\
\hline

ACMMP & 71.57 & 83.42 & 92.03 & 95.54 & 97.37\\
w/. NESP & 74.54 & 85.33 & 93.25 & 96.45 & 97.99\\
w/. ES (ours) & \textbf{76.01} & \textbf{86.60} & \textbf{94.07} & \textbf{97.07} & \textbf{98.37} \\
\hline
\end{tabular}
% }
\caption{Comparison of F$_1$-scores of different sampling strategies on the ETH3D training dataset.}
\label{table4}
\end{table}

\subsection{Ablation Studies}

In the ETH3D training dataset, we conduct ablation experiments to verify the effectiveness of each component in our proposed method, which includes Edge-Guided Non-Local Sampling (ES), Perceptual Range Expansion (PE), Plane Construction Optimization (PO), and Adaptive Patch Size Adjustment (AA). Tab. 3 illustrates the effectiveness of each part. ES significantly expands the solution space for reliable pixel hypotheses, enhancing their accuracy and improving plane construction for unreliable pixels. PE, similar to ES, extends the search range for reliable pixels within adaptive patch deformation, helping to avoid the influence of locally optimal reliable pixels during plane model construction. PO optimizes the RANSAC algorithm, ensuring the quality of the final plane models. Together, ES, PE, and PO substantially improve completeness. AA adjusts the center patch size, enabling faster convergence and allowing hypotheses close to the correct solution to be refined with greater precision within limited iterations, thereby enhancing accuracy.

We also compare ES with the NESP module of HPM-MVS, integrating ES into ACMM, ACMP, and ACMMP, following \cite{Ren2023}. As shown in Tab. 4, our method outperforms others, except at the 1cm and 2cm thresholds in ACMM. The lack of plane priors in ACMM leads to erroneous depth estimations in low-textured areas, making it less suitable for our broader sampling range compared to NESP.

Further results are provided in the supplementary materials, including additional experimental details and comparative studies, extensive point cloud visualizations.

\section{Conclusion}
In this paper, we propose the DPE-MVS method, which addresses issues in plane construction by introducing dual-level precision edge information. Our method significantly improves reconstruction of low-textured areas and excels in recovering stochastic textured areas.  Experimental results demonstrate SOTA performance on ETH3D and Tanks \& Temples. However, improvement is needed in handling fine scene details, where learning-based methods excel. Future work will integrate these methods for more accurate results.
% However, improvement is needed in handling fine scene details, an area where learning-based methods excel. 
% Future work will integrate learning-based methods for more comprehensive and accurate results.

\section{ Acknowledgements}

This work was supported in part by the Strategic Priority Research Program of the Chinese Academy of Sciences under Grant No.XDA0450203, in part by the Program of National Natural Science Foundation of China under Grant 62172392.

% \cite{Bleyer2011} PatchMatchStereo
% \cite{Cremers2010} 基于表面迭代优化的方法
% \cite{Ding2022}  Transmvsnet
% \cite{Furukawa2009} 基于Patch的方法
% \cite{Furukawa2015} Multi-view stereo: A tutorial
% \cite{Galliani2015} Gipuma
% \cite{Jensen2014} DTU
% \cite{Knapitsch2017} Tanks and Temples
% \cite{Li2024} DS-PMNet
% \cite{Liao2019} PLC
% \cite{Gu2020} CasMVSNet
% \cite{Ren2023} HPM-MVS
% \cite{Romanoni2019} TAPA-MVS
% \cite{Schoenberger2016} Colmap
% \cite{Schops2017} ETH3D
% \cite{Stathopoulou2023} A survey on conventional and learning-based methods for multi-view stereo
% \cite{Su2023} EPNet
% \cite{Sun2022} API-MVS
% \cite{Tian2023} HQP-MVS
% \cite{Vogiatzis2007} 基于体素的方法
% \cite{Wang2021} Patchmatchnet
% \cite{Wang2022} Itermvs
% \cite{Wang2023} APD-MVS
% \cite{Wei2021} Aa-rmvsnet
% \cite{Wu2024} GoMVS
% \cite{Xu2019} ACMM/ACMH
% \cite{Xu2020} ACMP
% \cite{Xu2020a} MARMVS
% \cite{Xu2022} ACMMP
% \cite{Yao2018} MVSNet
% \cite{Yao2020} BlendedMVS
% \cite{Yuan2024} SD-MVS
% \cite{Yuan2024a} TSAR-MVS
% \cite{Zhang2022} CPTT-MVS
% \cite{Zhang2023} N2MVSNet
% \cite{Zhou2021} DP-MVS
% \cite{kuhn2019plane} PCF-MVS

% \bibliography{aaai25.bib}

\section*{Supplementary Materials}

\subsection{Review of Conventional PM}

The conventional PatchMatch (PM) method, referred to as ACMH \cite{Xu2019}, involves four basic steps of PatchMatch: random initialization, propagation, multi-view matching cost calculation, and refinement. This section focuses on propagation and cost calculation.

\medskip
\noindent \textbf{Adaptive Checkerboard Sampling} Propagation refers to the process of hypothese sampling from surrounding pixels. ACMH partitions the pixels of the reference view into a red-black checkerboard pattern. This method allows for parallel hypothesis updating, where red pixels can be used to update the hypotheses of black pixels, and vice versa. Then, ACMH sets the sampling area for each pixel to four V-shaped areas and four long strip areas, with these regions being fixed. Finally, the hypothesis of the pixel with the smallest matching cost in each area is selected. The matching cost here refers to the calculation based on random initialization or the results from the previous iteration. This sampling strategy exhibits adaptability within local regions but is prone to local optima.

\medskip
\noindent \textbf{Matching Cost Calculation} For a given pixel $\mathbf{p}$ in the reference image $I_{ref}$, given a plane hypothesis $\boldsymbol{\theta}_p = [\mathbf{n}^T,d]^T$, the patch $\mathbf{B}_p$ centered at this pixel can be projected to to the patches $\mathbf{B}_p^j$ in the j-th source image $I_{src}^j$ through a homography transformation as follows:
\begin{equation}
    \mathbf{H}_j=\mathbf{K}_j\left(\mathbf{R}_j \mathbf{R}^{-1}_{ref}+\frac{\mathbf{R}_j(\mathbf{C}_{ref}-\mathbf{C}_j)\mathbf{n}^T}{\mathbf{n}^Td\mathbf{K}^{-1}_{ref}\mathbf{p}}\mathbf{K}^{-1}_{ref}\right),
\end{equation}
where $\mathbf{K}$ denotes the intrinsic matrix, $\mathbf{R}$ represents the rotation matrix, and $\mathbf{C}$ indicates the camera center.
The similarity between two patches is measured as follows: 

\begin{equation}
    NCC(\mathbf{B}_p, \mathbf{B}_p^j) = \frac{cov (\mathbf{B}_p, \mathbf{B}_p^j)}{\sqrt{cov (\mathbf{B}_p, \mathbf{B}_p),cov (\mathbf{B}_p^j, \mathbf{B}_p^j)}},
\end{equation}
where $cov (\mathbf{X}, \mathbf{Y})$ represents the covariance of pixel intensities between two patches. The matching cost $m_j (\mathbf{p}, \boldsymbol{\theta}_p, \mathbf{B}_p)$ for the $I_{src}^j$ is defined as $1 - NCC(\mathbf{B}_p, \mathbf{B}_p^j)$.
The final matching cost is the weighted sum of the costs from the source images, as follows:

\begin{equation}
    m (\mathbf{p}, \boldsymbol{\theta}_p, \mathbf{B}_p) = \frac{\sum_j w_j \cdot m_j (\mathbf{p}, \boldsymbol{\theta}_p, \mathbf{B}_p)}{\sum_j w_j},
\end{equation}
where $w_j$ is the weight of $I_{src}^j$. The hypothesis with the smallest cost will be selected to update the current hypothesis. Due to the fixed size of $\mathbf{B}_p$, this approach faces matching ambiguity in stochastic textured and low-textured areas, where it is difficult to capture effective features.

\subsection{More Experiments and Analysis}

In this section, we present a more detailed description of the datasets utilized and include additional comparative experiments to further highlight the advantages of our method. All experiments were run on a Silver 4210 CPU and a GeForce RTX 3090 GPU with 24GB of VRAM.

\begin{table*}[htbp]
\centering
% \resizebox{.95\textwidth}{!}{
\fontsize{9pt}{10pt}\selectfont
\setlength{\tabcolsep}{1mm}{
\begin{tabular}{c c c c|c c c c c c c|c c c c c c|c}
\hline

\multicolumn{4}{c|}{Settings} & \multicolumn{7}{|c|}{indoor} & \multicolumn{6}{|c|}{outdoor} & \multirow{2}{*}{Ave.} \\
\cline{1-17}

ES & PE & PO & AA &delive. & kicker & office & pipes & relief & relief\_2 & terrains & courty. & electro & facade & meadow & playgr. & terrace & \\
\hline

& & & &  	
90.10 & 87.38 & 90.68 & 86.82 & 87.13 & 86.45 & 93.83 & 90.66 & 90.78 & 75.64 & 78.72 & 80.15 & 90.51 & 86.84\\

\ding{51} & & & & 91.85 & 87.32 & 90.82 & 88.82 & 87.81 & 86.83 & 94.60 & 91.11 & 91.32 & 75.87 & 76.31 & 79.23 & 90.94 & 87.14\\

& \ding{51} & & & 91.49 & 87.25 & 92.26 & 88.98 & 87.59 & 86.58 & 94.82 & 90.80 & 90.85 & 75.58 & 76.78 & 79.16 & 90.67 & 87.14\\

\ding{51} & \ding{51} & & & 91.89 & 87.87 & 92.48 & 89.56 & 88.20 & 86.94 & 95.08 & \underline{91.15} & 91.35 & 75.85 & 77.78 & 80.11 & 90.85 & 87.62\\

\ding{51} & \ding{51} & \ding{51} & & \textbf{92.27} & 89.05 & \underline{93.70} & \underline{90.26} & \textbf{88.75} & \underline{87.57} & \underline{95.35} & 90.74 & \underline{91.96} & \underline{75.94} & 77.79 & 80.27 & \textbf{90.98} & 88.05\\

\ding{51} & \ding{51} & & \ding{51} & 91.87 & \underline{89.33} & 92.53 & 89.95 & 88.29 & 87.03 & 95.19 & \textbf{91.22} & 91.53 & 75.91 & \underline{80.54} & \underline{80.71} & 90.87 & \underline{88.08}\\

\ding{51} & \ding{51} & \ding{51} & \ding{51} & \underline{92.20} & \textbf{90.55} & \textbf{93.85} & \textbf{90.75} & 
\underline{88.62} & \textbf{87.59} & \textbf{95.48} & 90.85 & \textbf{91.98} & \textbf{76.00} & \textbf{82.09} & \textbf{81.25} & \underline{90.96} & \textbf{88.63}\\
\hline 

\end{tabular}
}
\caption{F$_1$-scores of our method with different settings on the ETH3D training dataset at a 2cm threshold; baseline: APD-MVS.}
\label{table5}
\end{table*}

\medskip
\noindent \textbf{Brief Review of Datasets} The ETH3D dataset \cite{Schops2017} and the Tanks \& Temples dataset \cite{Knapitsch2017} are widely used in MVS (multi-view stereo) tasks. The high-resolution multi-view benchmark in the ETH3D dataset is characterized by significant viewpoint changes. This dataset, with an image resolution of approximately $6048 \times 4032$ resolution, covers diverse indoor and outdoor scenes, enabling a comprehensive evaluation of algorithms under high-resolution and multi-view conditions. The Tanks \& Temples dataset is divided into Intermediate and Advanced sets based on scene scale, with an image resolution of approximately $1920 \times 1080$ resolution. This dataset focuses on complex real-world scenes, providing continuous video frame images, and is commonly used to assess the generalization capability of MVS methods.

ETH3D assesses MVS methods using accuracy, completeness, and F$_1$-score. Accuracy measures the average distance between reconstructed points and the ground truth, while completeness evaluates how much of the ground truth is captured. The F$_1$-score combines both metrics to provide an overall performance measure.

Tanks \& Temples evaluates MVS methods with precision, recall, and F$_1$-score. Precision measures the alignment of reconstructed points with the ground truth, recall assesses the coverage of ground truth points, and F1-score balances both to indicate overall performance.

\medskip
\noindent\textbf{Further Results Presentation} As shown in Tab. 5, we present additional results from the ablation study, specifically the F$_1$-scores for each scene in the ETH3D training dataset at a threshold of 2cm. The settings include Edge-Guided Non-Local Sampling (ES), Perceptual Range Expansion (PE), Plane Construction Optimization (PO), and Adaptive Patch Size Adjustment (AA). The implementation of ES has led to significant improvements in most scenes. PE and PO have proven effective in scenes with large, low-texture planes, such as office and pipes, by enhancing corresponding metrics. AA effectively handles stochastic textured regions, as demonstrated by its performance in the meadow scene, which includes two houses and a large lawn. Similarly, the improvements in the office and pipes scenes highlight AA's effectiveness in dealing with low-textured areas.

As illustrated in Fig. 8, our method achieves high recall on the Tanks \& Temples dataset, with M60 from the Intermediate dataset and Courtroom from the Advanced dataset. These results demonstrate the robustness and effectiveness of our method.

\begin{figure}[t]
\centering
\includegraphics[width=1\columnwidth]{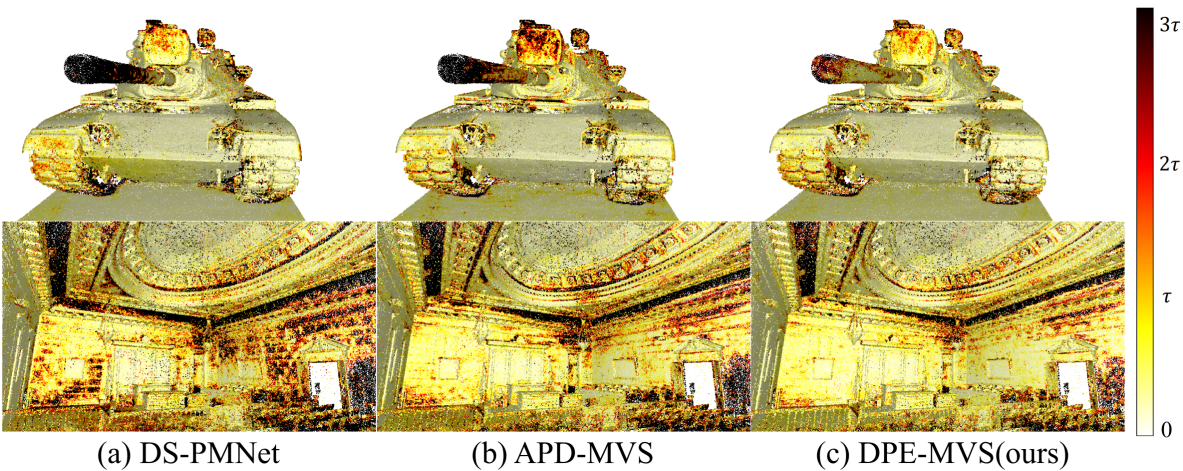} % Reduce the figure size so that it is slightly narrower than the column. Don't use precise values for figure width.This setup will avoid overfull boxes.
\caption{Qualitative comparison of recall. We compared our method with DS-PMNet \cite{Li2024} and APD-MVS on M60 and Courtroom, lighter areas indicate better reconstruction. Our method demonstrated the best performance.}
\label{fig8}
\end{figure}

\medskip
\noindent\textbf{Memory and Runtime Analysis} We compare the memory usage and runtime between APD-MVS and our proposed method when processing an image with a resolution of $6221 \times 4146$. Compared to APD-MVS, our method requires additional memory to store edge information extracted from the reference image. As $6221 \times 4146$ represents nearly the maximum resolution in the datasets, our method requires up to 2GB of additional memory, as shown in Tab. 6.

For runtime comparison, we incorporated an edge extraction preprocessing step, enabling efficient processing of high-resolution images using only the CPU, as shown in Tab. 7. A full PM iteration includes adaptive patch deformation, three iterations of conventional PM, three iterations of deformable PM, and additional operations not explicitly detailed. In the adaptive patch deformation stage, we introduced operations to achieve a more accurate plane model, which increased processing time by 6.1 seconds. During the conventional PM stage, we expanded the search range, resulting in only a slight runtime increase of 0.29 seconds. In the deformable PM stage, we adjusted the size of the center patch: when the center patch size was increased, the number of pixels to be matched remained constant, and when the center patch was discarded, the matching process accelerated. This adjustment resulted in a time savings of 5.72 seconds. As a result, the total runtime for a full PM iteration is nearly the same as the original method, demonstrating that our method achieves significant improvements with minimal impact on computational efficiency.

\begin{table}[t]
\centering
% \resizebox{.95\textwidth}{!}{
\centering
% \resizebox{.95\textwidth}{!}{
\fontsize{9pt}{10pt}\selectfont
\begin{tabular}{c | c}
\hline

Method & GPU Mem. (GB)\\
\hline
APD-MVS & 6.8\\
DPE-MVS & 8.8\\
\hline

\end{tabular}
% }
\caption{GPU Memory comparison between APD-MVS and DPE-MVS for a $6221 \times 4146$ image. }
\label{table6}
\end{table}

\begin{table}[t]
\centering
% \resizebox{.95\textwidth}{!}{
\centering
% \resizebox{.95\textwidth}{!}{
\fontsize{9pt}{11pt}\selectfont
\begin{tabular}{c | c c}
\hline

{\multirow{2}*{Stages}} & \multicolumn{2}{|c}{Time Cost (s)}\\
\cline{2-3}
& APD-MVS & DPE-MVS \\
\hline
Extract Fine Edge & - & 0.08\\
Extract Corse Edge & - & 2.83\\

\hline

Adaptive Patch Deformation & 2.94 & 9.04\\
Conventional PM Iteration & 0.90 & 1.19 \\
Deformable PM Iteration & 23.33 & 17.61 \\
\hline
Full PM Iteration & 100.89 & 103.56\\
\hline
\end{tabular}
% }
\caption{Runtime comparison between APD-MVS and DPE-MVS for a $6221 \times 4146$ image. }
\end{table}

\subsection{Additional Point Cloud Results} 

This section supplements the point cloud results obtained by our method. Fig. 9 and Fig. 10 present the point cloud results of the Intermediate and Advanced datasets, respectively. Fig. 11 and Fig. 12 show the point cloud results of the ETH3D training and test datasets, respectively.

\clearpage

\begin{figure*}[tb]
 \centering
 \begin{minipage}{.1\textwidth}
    \centerline{Family}
 \end{minipage}%
 \begin{minipage}{.65\linewidth}
%  	\vspace{3pt}
    \centerline{\includegraphics[width=\textwidth,trim=10 10 10 10,clip]{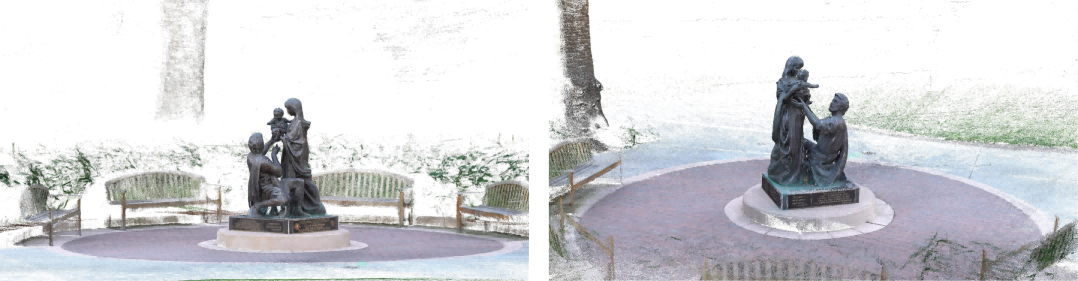}}
    \vspace{3pt}
 \end{minipage}
 
 \begin{minipage}{.1\textwidth}
    \centerline{Francis}
 \end{minipage}%
 \begin{minipage}{.65\linewidth}
% 	\vspace{3pt}
	\centerline{\includegraphics[width=\textwidth,trim=10 10 10 10,clip]{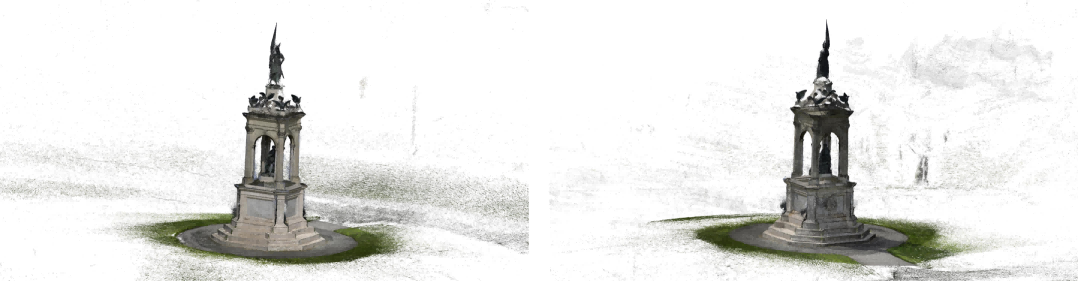}}
	\vspace{3pt}
\end{minipage}

 \begin{minipage}{.1\textwidth}
    \centerline{Horse}
 \end{minipage}%
 \begin{minipage}{.65\linewidth}
%  	\vspace{3pt}
 	\centerline{\includegraphics[width=\textwidth,trim=10 10 10 10,clip]{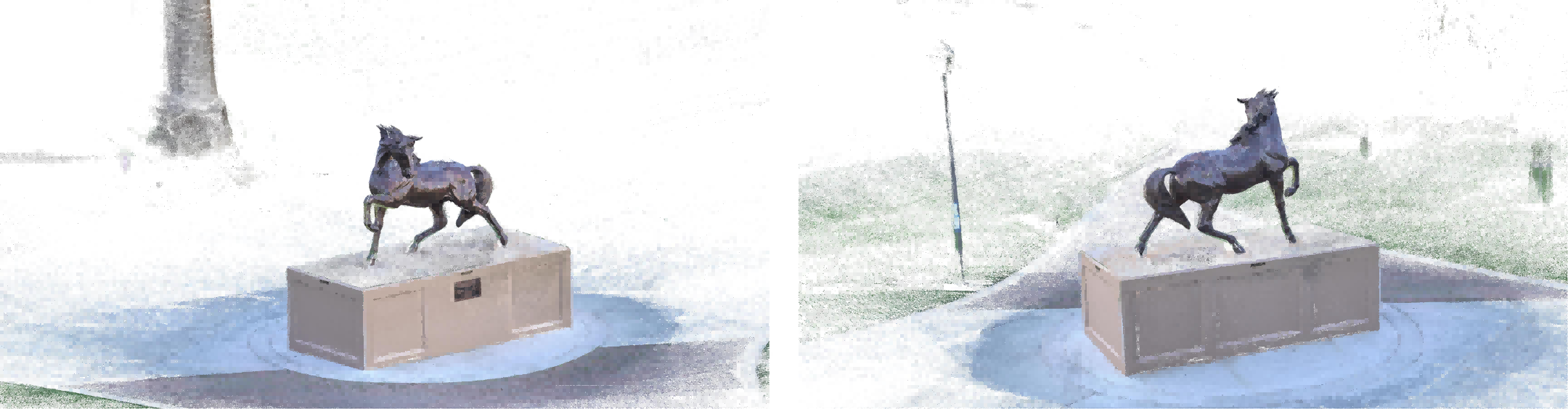}}
 	\vspace{3pt}
 \end{minipage}
 
 \begin{minipage}{.1\textwidth}
    \centerline{Lighthouse}
 \end{minipage}%
 \begin{minipage}{.65\linewidth}
% 	\vspace{3pt}
	\centerline{\includegraphics[width=\textwidth,trim=10 10 10 10,clip]{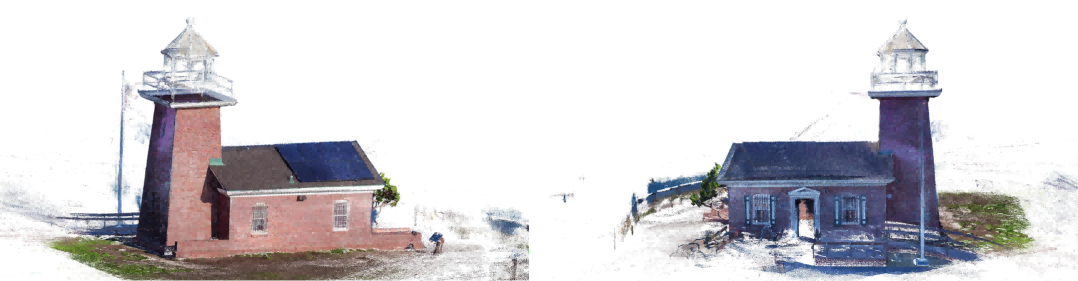}}
	\vspace{3pt}
\end{minipage}
 
 \begin{minipage}{.1\textwidth}
    \centerline{M60}
 \end{minipage}%
 \begin{minipage}{.65\linewidth}
%  	\vspace{3pt}
 	\centerline{\includegraphics[width=\textwidth,trim=10 10 10 10,clip]{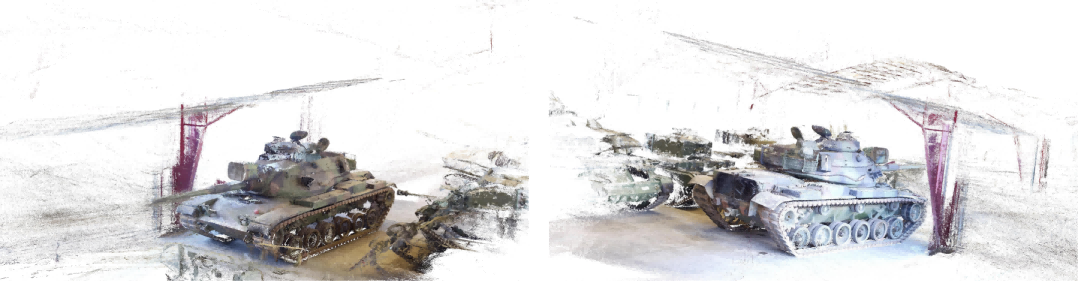}}
 	\vspace{3pt}
 \end{minipage}
 
 \begin{minipage}{.1\textwidth}
    \centerline{Panther}
 \end{minipage}%
 \begin{minipage}{.65\linewidth}
% 	\vspace{3pt}
	\centerline{\includegraphics[width=\textwidth,trim=10 10 10 10,clip]{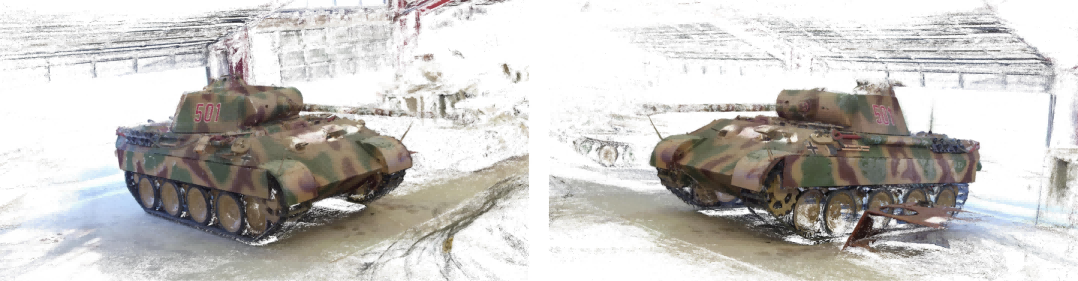}}
	\vspace{3pt}
\end{minipage}

 \begin{minipage}{.1\textwidth}
    \centerline{Playground}
 \end{minipage}%
 \begin{minipage}{.65\linewidth}
%  	\vspace{3pt}
 	\centerline{\includegraphics[width=\textwidth,trim=10 10 10 10,clip]{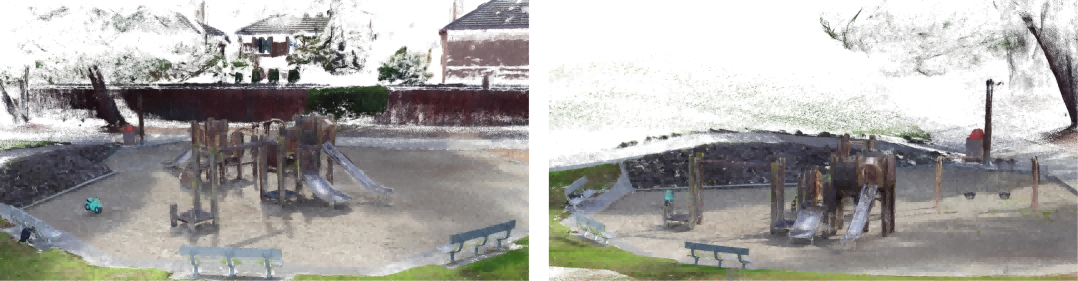}}
 	\vspace{3pt}
 \end{minipage}
 
 \begin{minipage}{.1\textwidth}
    \centerline{Train}
 \end{minipage}%
 \begin{minipage}{.65\linewidth}
% 	\vspace{3pt}
	\centerline{\includegraphics[width=\textwidth,trim=10 10 10 10,clip]{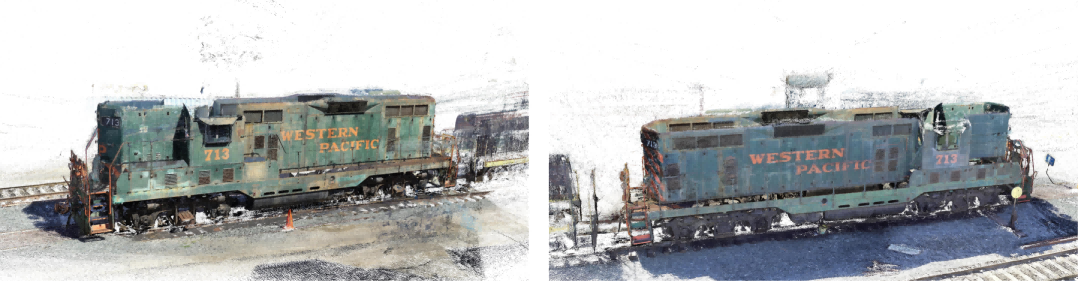}}
	\vspace{3pt}

\end{minipage}

\caption{Point cloud results of the Intermediate dataset.}
\label{fig9}

\end{figure*}

\begin{figure*}[tb]
 \centering

 \begin{minipage}{.1\textwidth}
    \centerline{Auditorium}
 \end{minipage}%
 \begin{minipage}{.7\linewidth}
 	\vspace{3pt}
 	\centerline{\includegraphics[width=\textwidth,trim=10 10 10 10,clip]{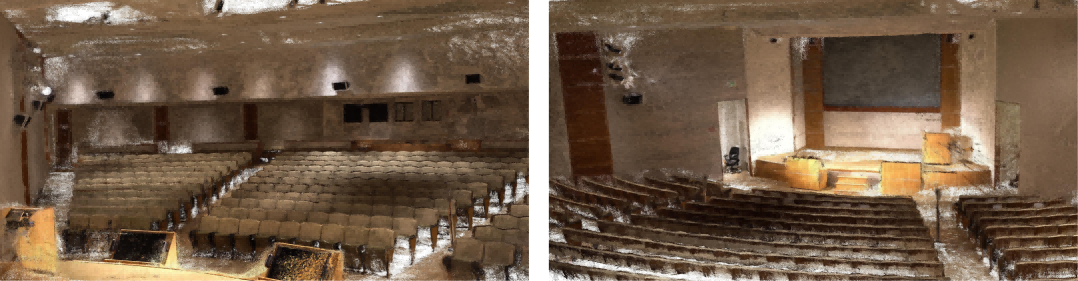}}
 	\vspace{3pt}
 \end{minipage}
 
 \begin{minipage}{.1\textwidth}
    \centerline{Ballroom}
 \end{minipage}%
 \begin{minipage}{.7\linewidth}
	\vspace{3pt}
	\centerline{\includegraphics[width=\textwidth,trim=10 10 10 10,clip]{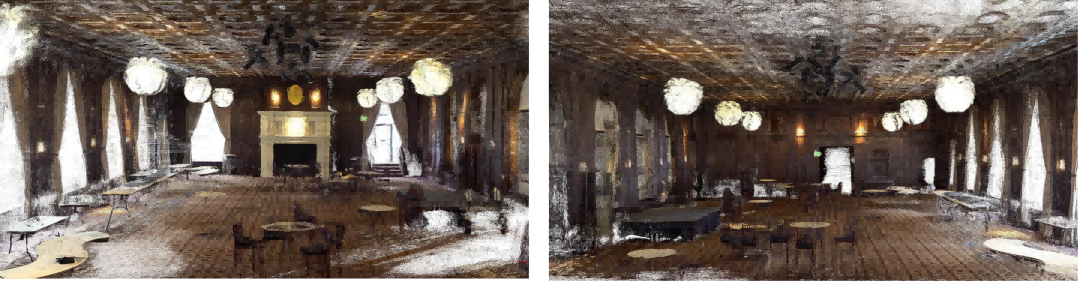}}
	\vspace{3pt}
\end{minipage}
 
 \begin{minipage}{.1\textwidth}
    \centerline{Courtroom}
 \end{minipage}%
 \begin{minipage}{.7\linewidth}
 	\vspace{3pt}
 	\centerline{\includegraphics[width=\textwidth,trim=10 10 10 10,clip]{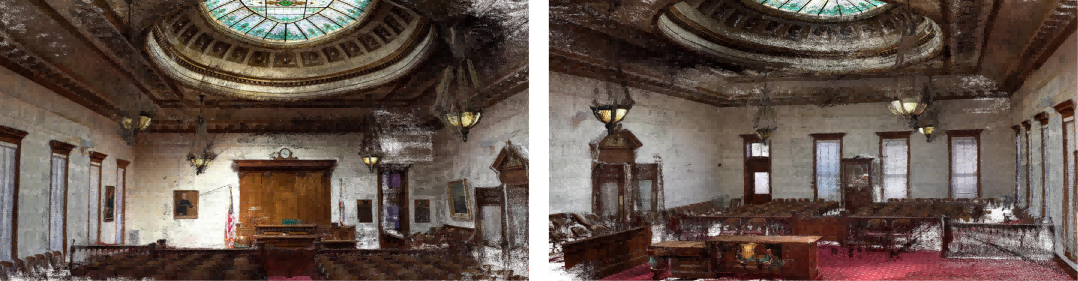}}
 	\vspace{3pt}
 \end{minipage}
 
 \begin{minipage}{.1\textwidth}
    \centerline{Museum}
 \end{minipage}%
 \begin{minipage}{.7\linewidth}
	\vspace{3pt}
	\centerline{\includegraphics[width=\textwidth,trim=10 10 10 10,clip]{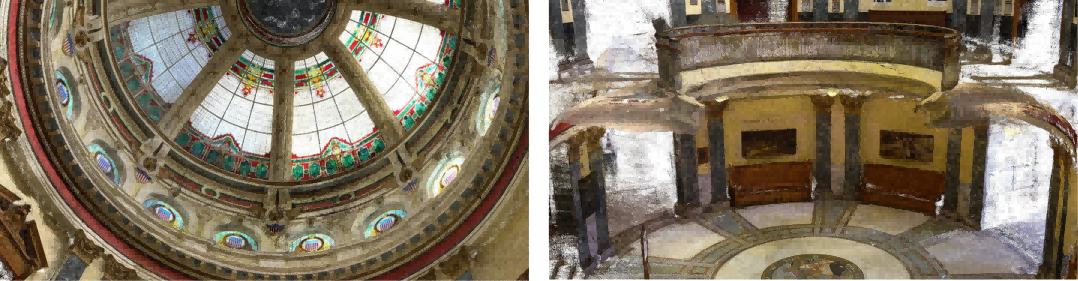}}
	\vspace{3pt}
\end{minipage}

 \begin{minipage}{.1\textwidth}
    \centerline{Palace}
 \end{minipage}%
 \begin{minipage}{.7\linewidth}
 	\vspace{3pt}
 	\centerline{\includegraphics[width=\textwidth,trim=10 10 10 10,clip]{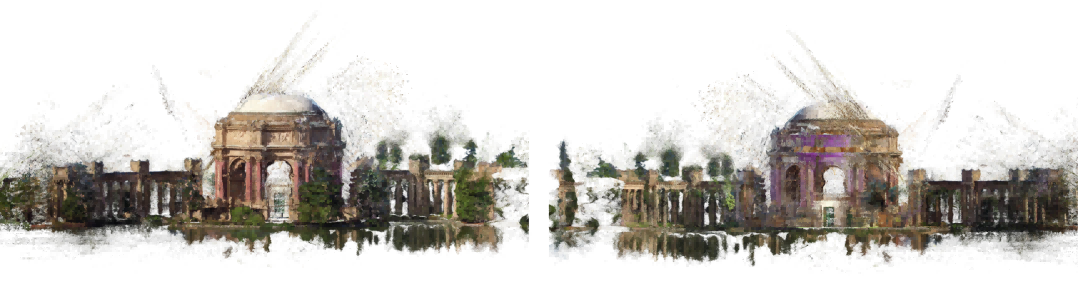}}
 	\vspace{3pt}
 \end{minipage}
 
 \begin{minipage}{.1\textwidth}
    \centerline{Temple}
 \end{minipage}%
 \begin{minipage}{.7\linewidth}
	\vspace{3pt}
	\centerline{\includegraphics[width=\textwidth,trim=10 10 10 10,clip]{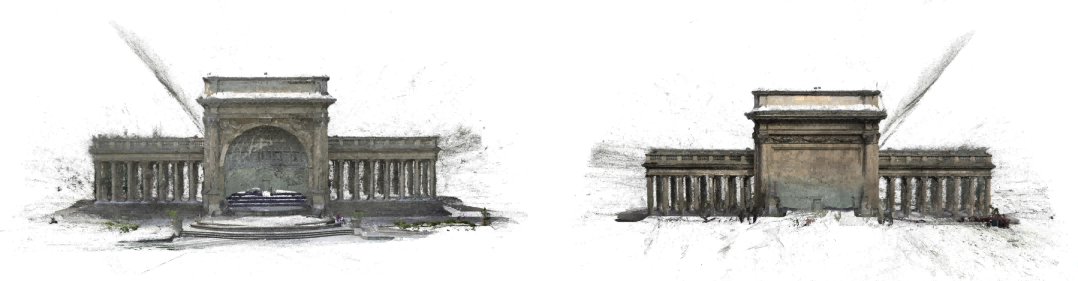}}
	\vspace{3pt}

\end{minipage}

\caption{Point cloud results of the Advanced dataset.}
\label{fig10}

\end{figure*}

\begin{figure*}[tb]
 \centering

 \begin{minipage}{.1\textwidth}
    \centerline{Courtyard}
    \centerline{\&}
    \centerline{Delivery}
 \end{minipage}%
 \begin{minipage}{.35\linewidth}
 	\vspace{3pt}
 	\centerline{\includegraphics[width=\textwidth]{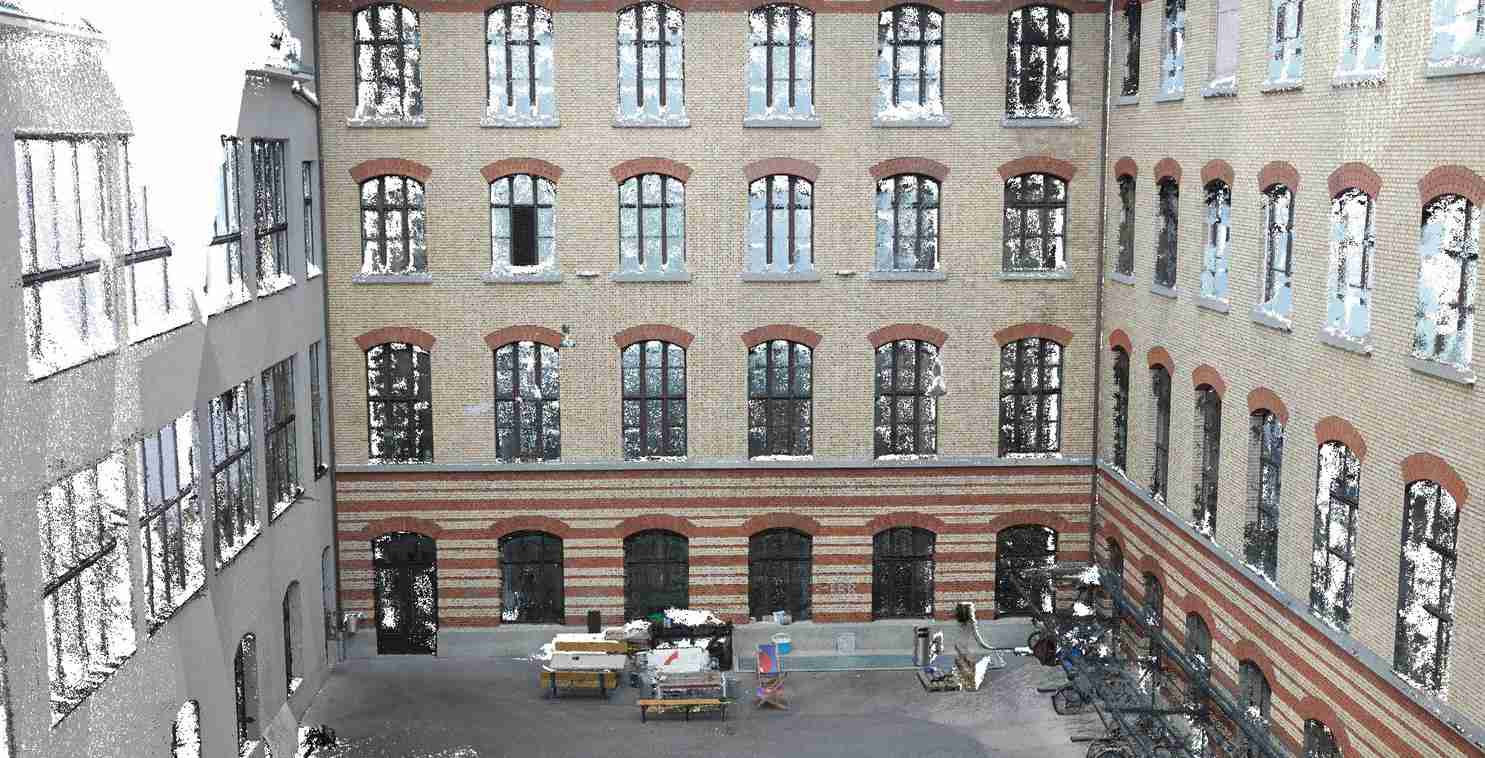}}
 	\vspace{3pt}
 \end{minipage}
  \begin{minipage}{.35\linewidth}
 	\vspace{3pt}
 	\centerline{\includegraphics[width=\textwidth]{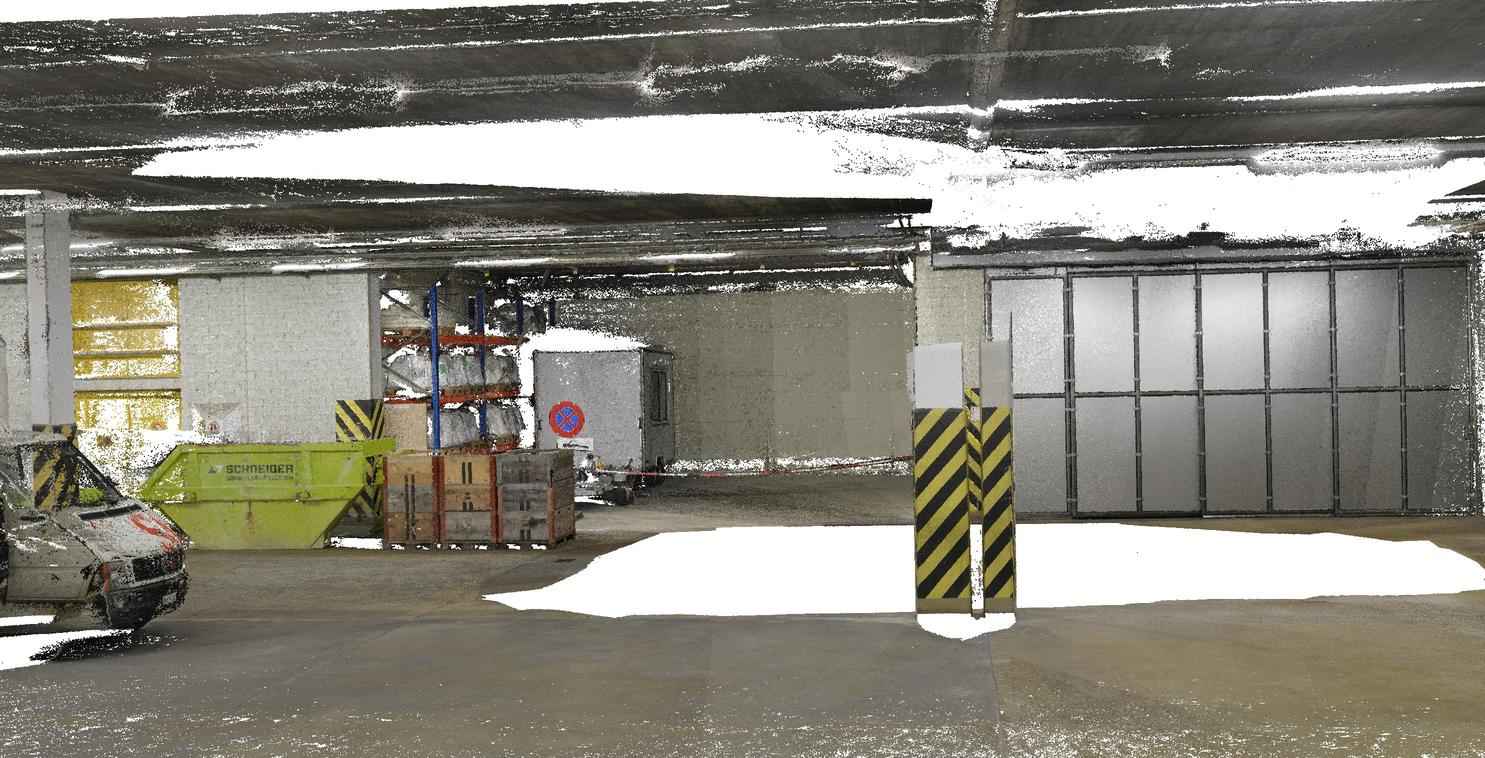}}
 	\vspace{3pt}
 \end{minipage}
 
  \begin{minipage}{.1\textwidth}
    \centerline{Electro}
    \centerline{\&}
    \centerline{Facade}
 \end{minipage}%
  \begin{minipage}{.35\linewidth}
 	\vspace{3pt}
 	\centerline{\includegraphics[width=\textwidth]{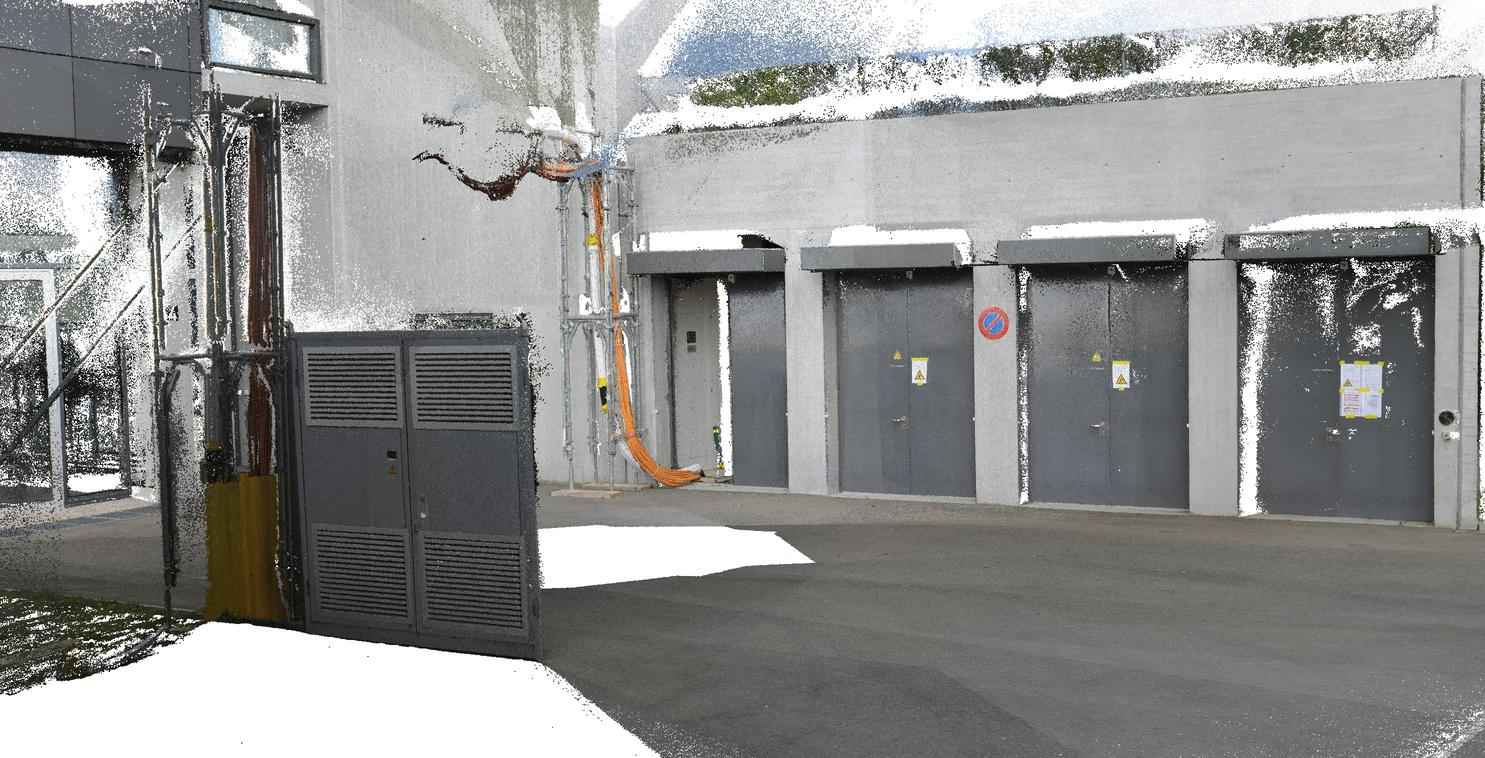}}
 	\vspace{3pt}
 \end{minipage}
  \begin{minipage}{.35\linewidth}
 	\vspace{3pt}
 	\centerline{\includegraphics[width=\textwidth]{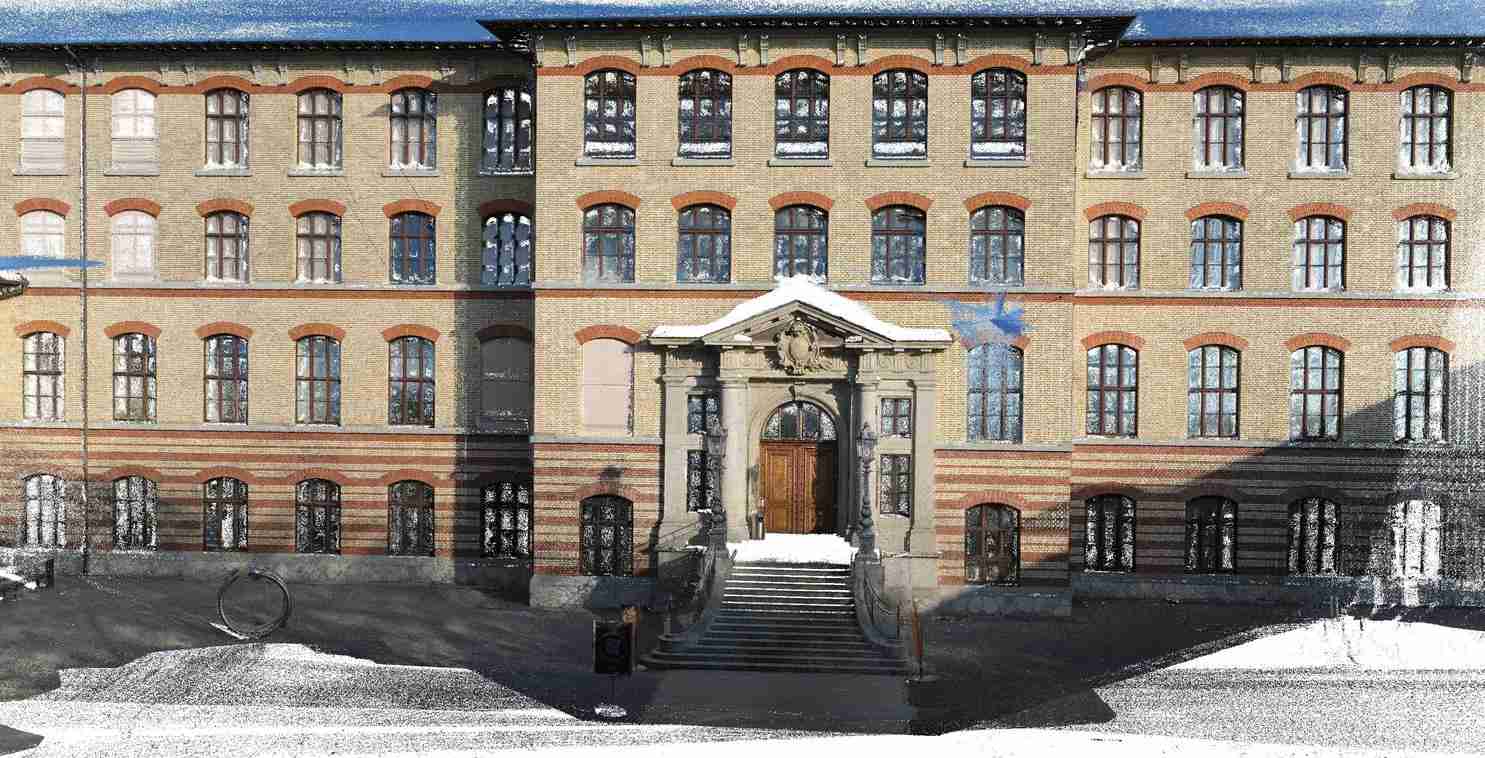}}
 	\vspace{3pt}
 \end{minipage}
 
 \begin{minipage}{.1\textwidth}
    \centerline{Kicker}
    \centerline{\&}
    \centerline{Meadow}
 \end{minipage}%
  \begin{minipage}{.35\linewidth}
 	\vspace{3pt}
 	\centerline{\includegraphics[width=\textwidth]{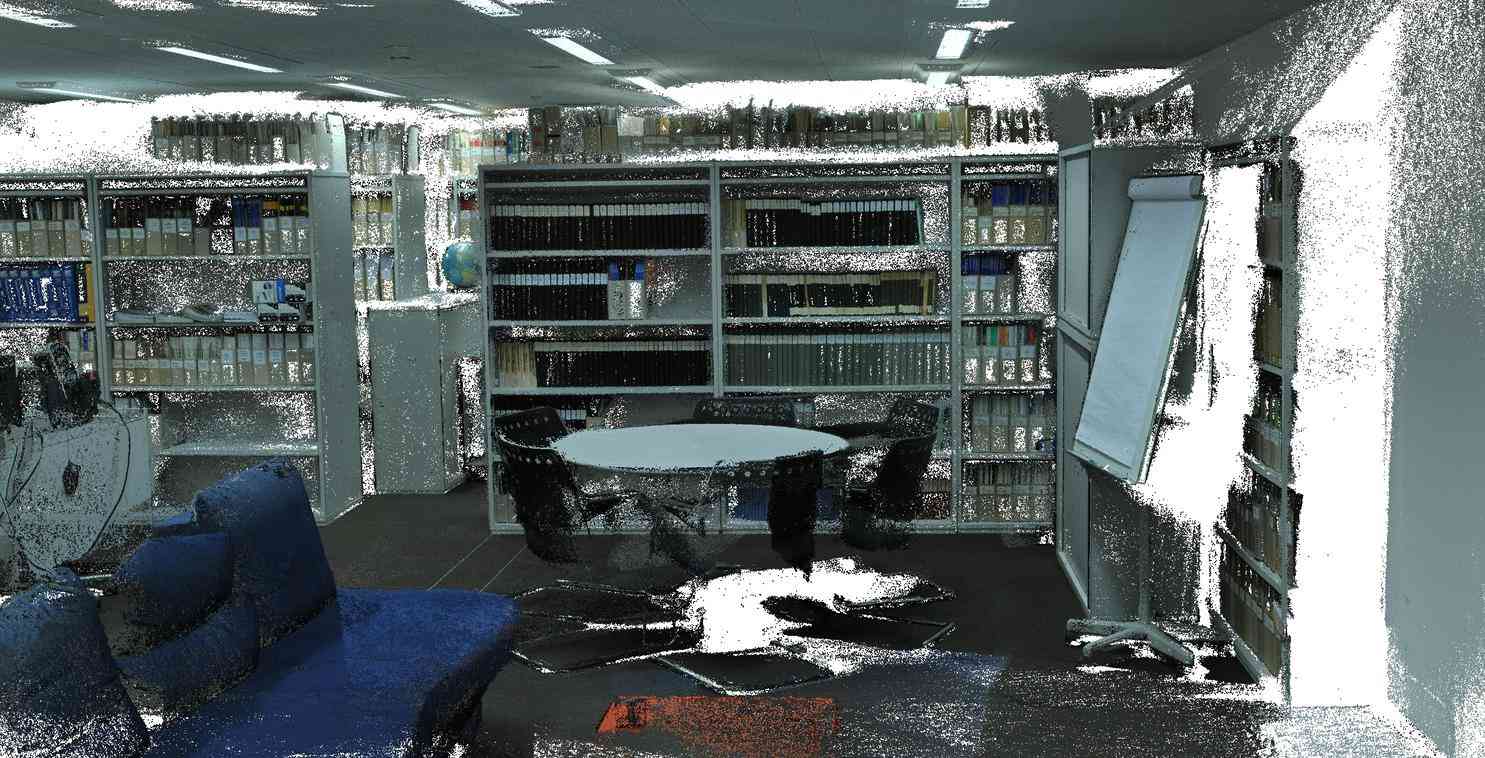}}
 	\vspace{3pt}
 \end{minipage}
  \begin{minipage}{.35\linewidth}
 	\vspace{3pt}
 	\centerline{\includegraphics[width=\textwidth]{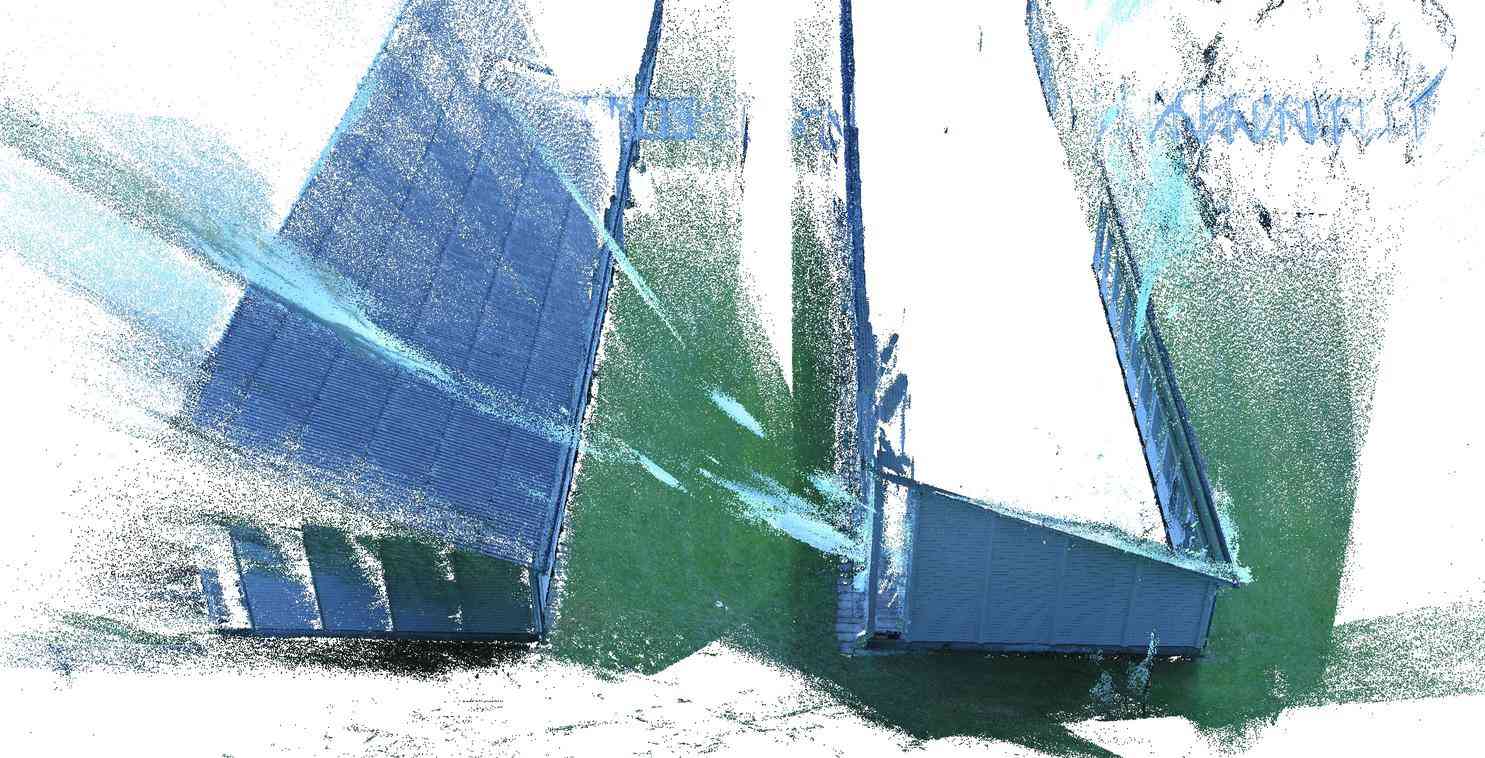}}
 	\vspace{3pt}
 \end{minipage}
 
  \begin{minipage}{.1\textwidth}
    \centerline{Office}
    \centerline{\&}
    \centerline{Pipes}
 \end{minipage}%
  \begin{minipage}{.35\linewidth}
 	\vspace{3pt}
 	\centerline{\includegraphics[width=\textwidth]{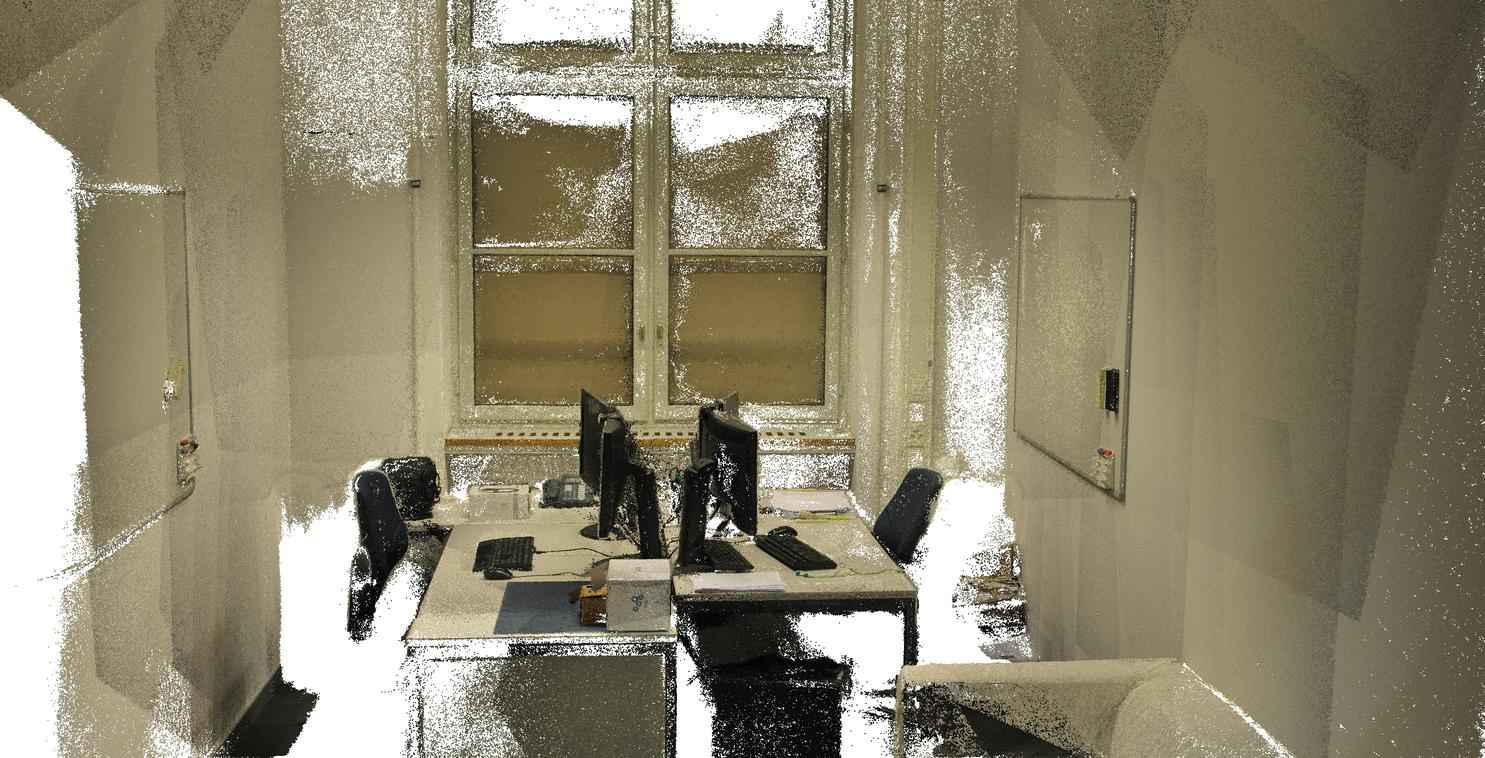}}
 	\vspace{3pt}
 \end{minipage}
  \begin{minipage}{.35\linewidth}
 	\vspace{3pt}
 	\centerline{\includegraphics[width=\textwidth]{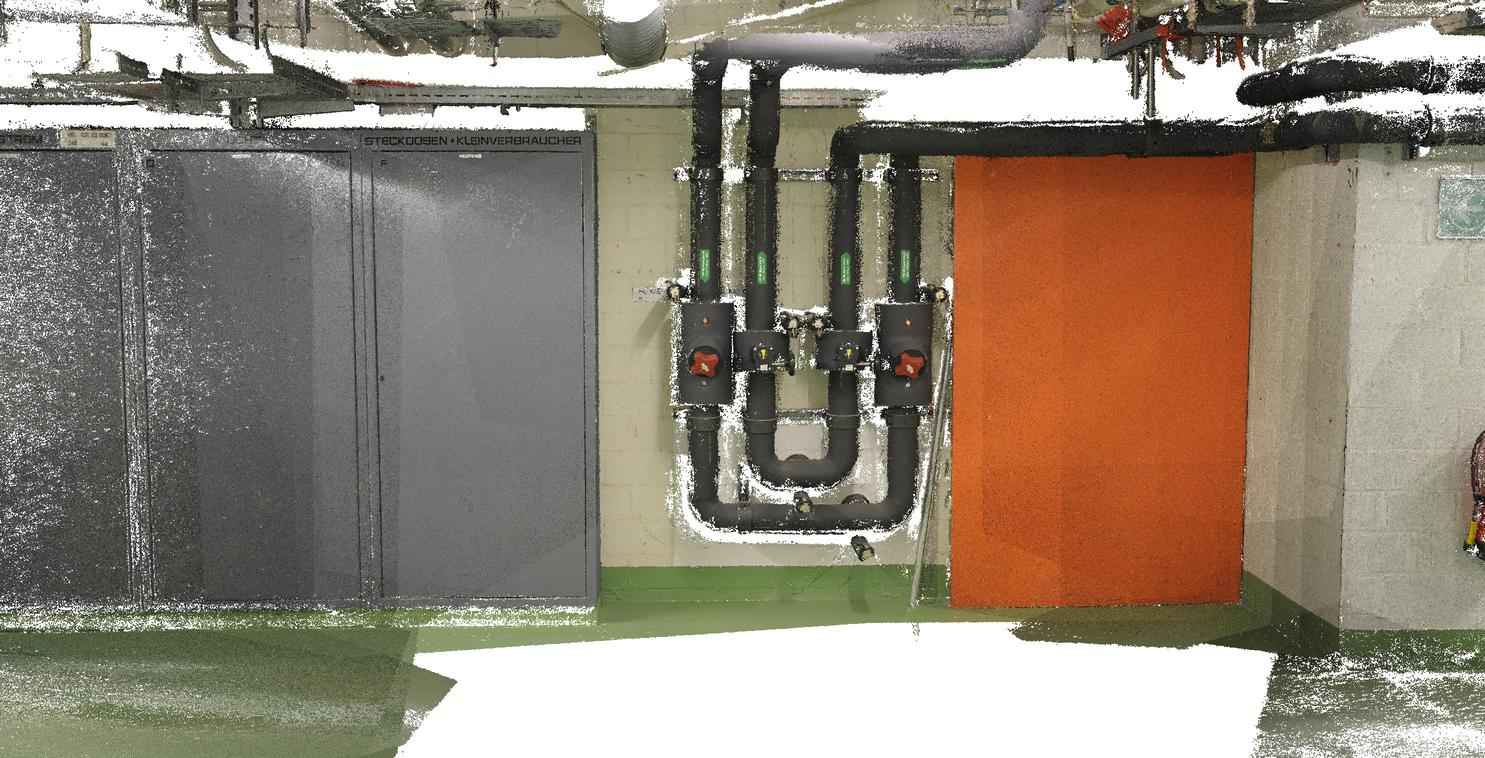}}
 	\vspace{3pt}
 \end{minipage}
 
  \begin{minipage}{.1\textwidth}
    \centerline{Playground}
    \centerline{\&}
    \centerline{Relief}
    \centerline{\&}
    \centerline{Relief 2}
 \end{minipage}%
  \begin{minipage}{.35\linewidth}
 	\vspace{3pt}
 	\centerline{\includegraphics[width=\textwidth]{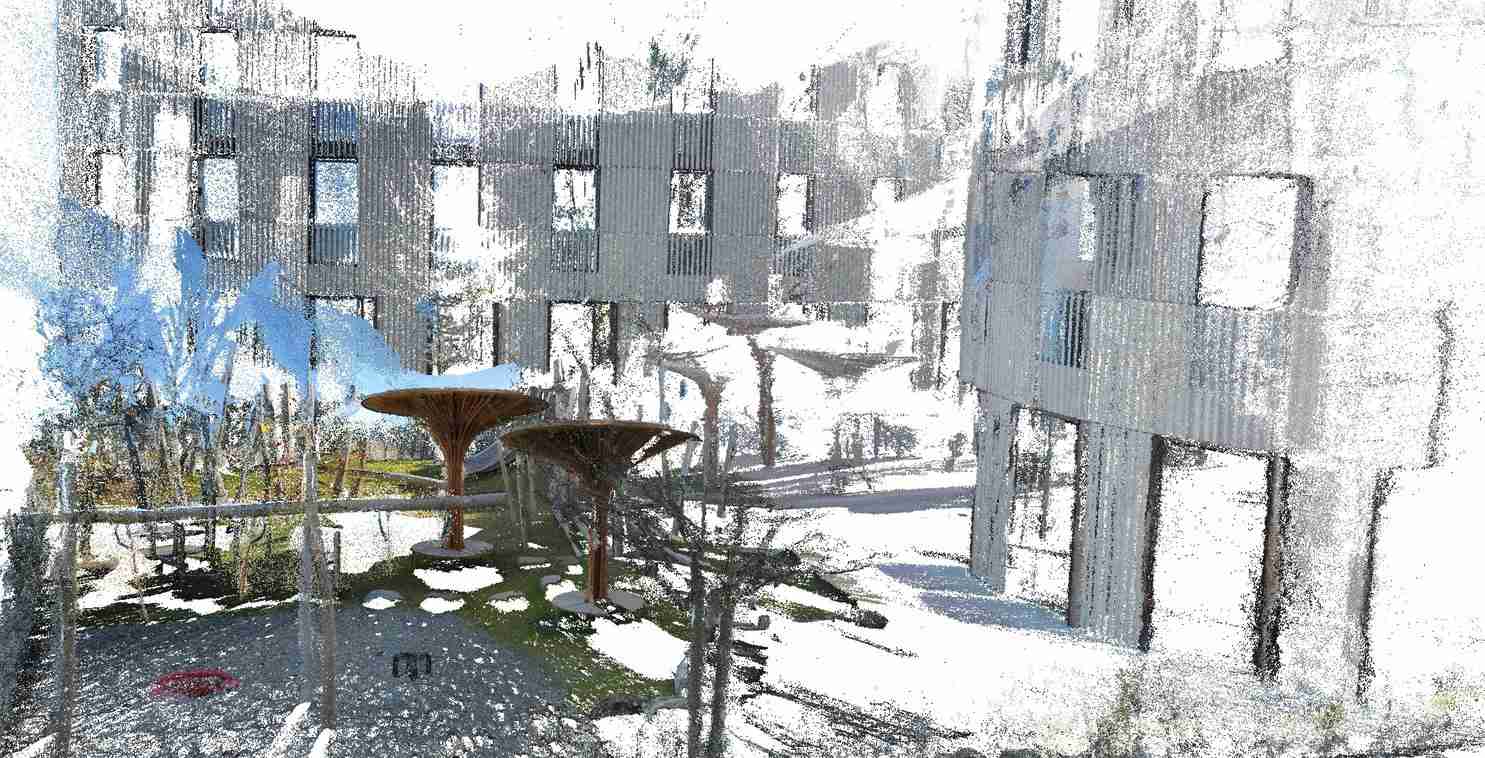}}
 	\vspace{3pt}
 \end{minipage}
 \begin{minipage}{.173\linewidth}
 	\vspace{3pt}
 	\centerline{\includegraphics[width=\textwidth]{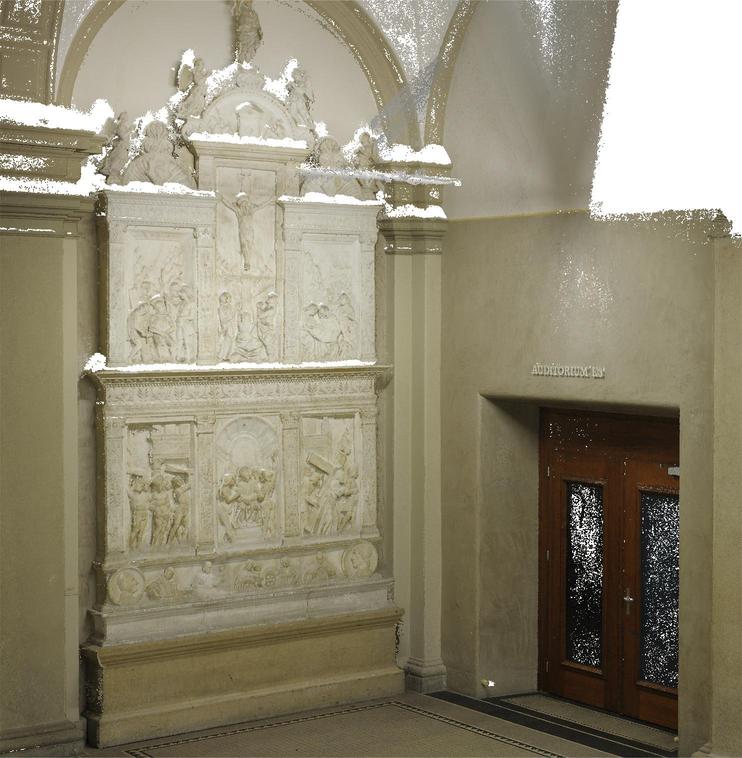}}
 	\vspace{3pt}
 \end{minipage}
 \begin{minipage}{.173\linewidth}
 	\vspace{3pt}
 	\centerline{\includegraphics[width=\textwidth]{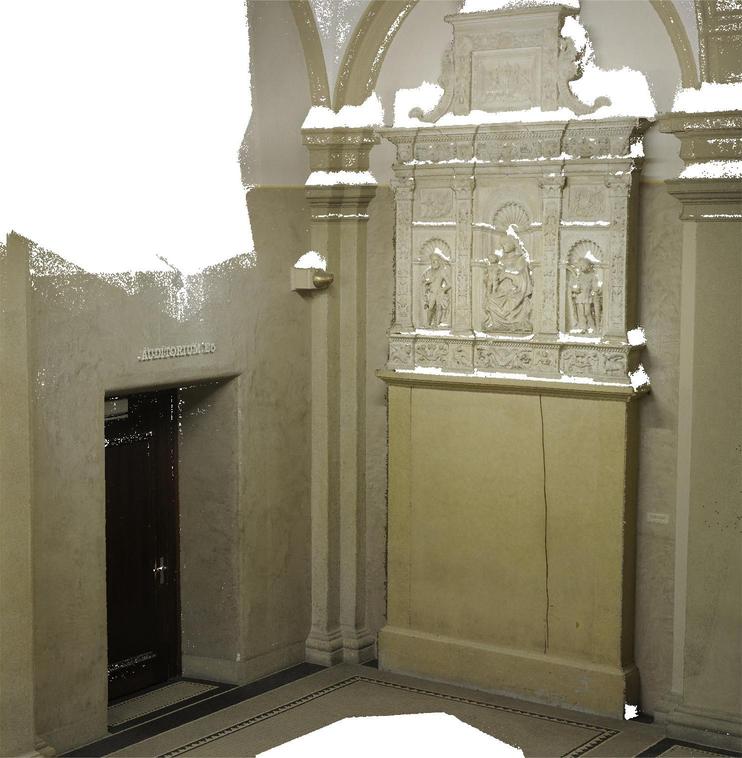}}
 	\vspace{3pt}
 \end{minipage}
 
 \begin{minipage}{.1\textwidth}
    \centerline{Terrace}
    \centerline{\&}
    \centerline{Terrains}
 \end{minipage}%
 \begin{minipage}{.35\linewidth}
 	\vspace{3pt}
 	\centerline{\includegraphics[width=\textwidth]{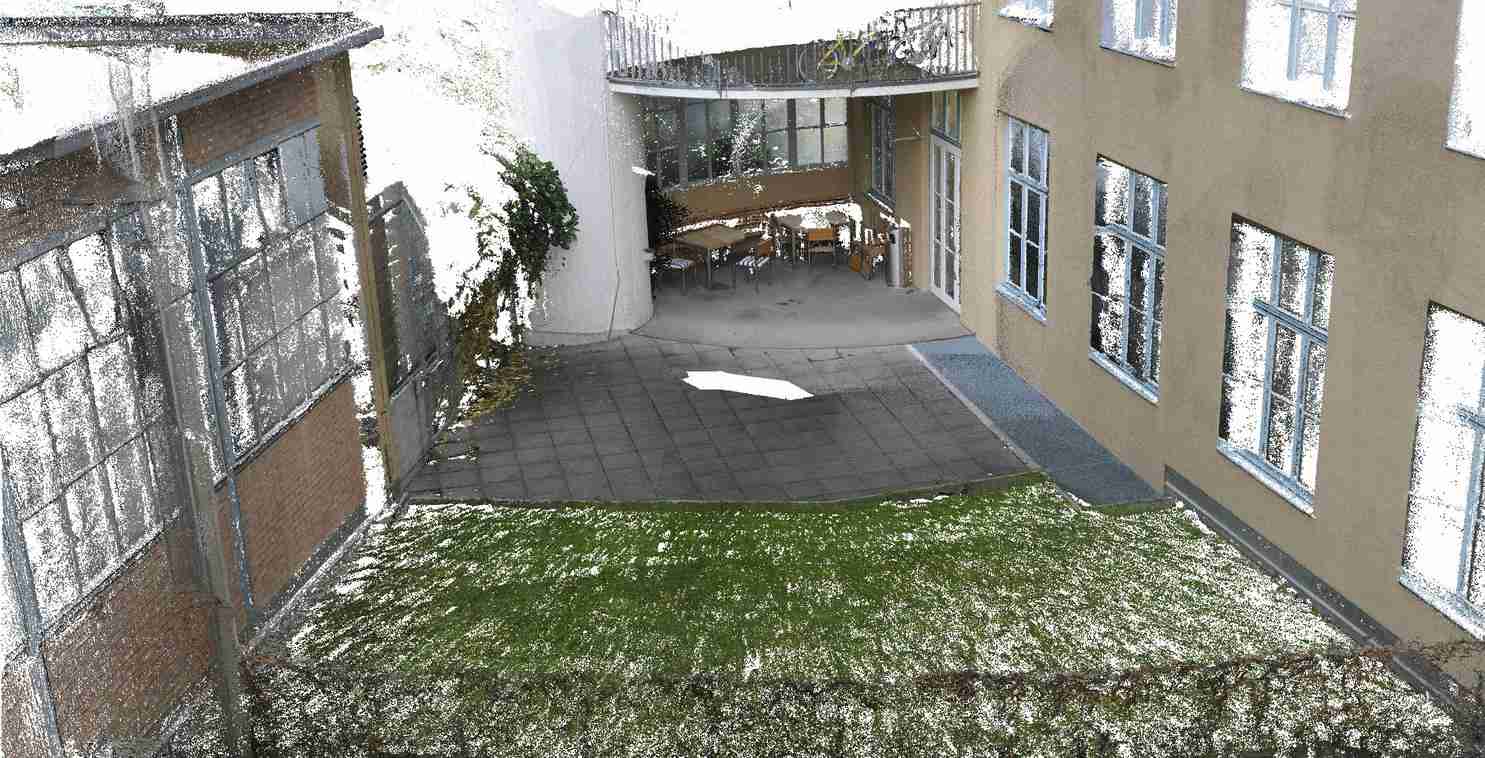}}
 	\vspace{3pt}
 \end{minipage}
  \begin{minipage}{.35\linewidth}
 	\vspace{3pt}
 	\centerline{\includegraphics[width=\textwidth]{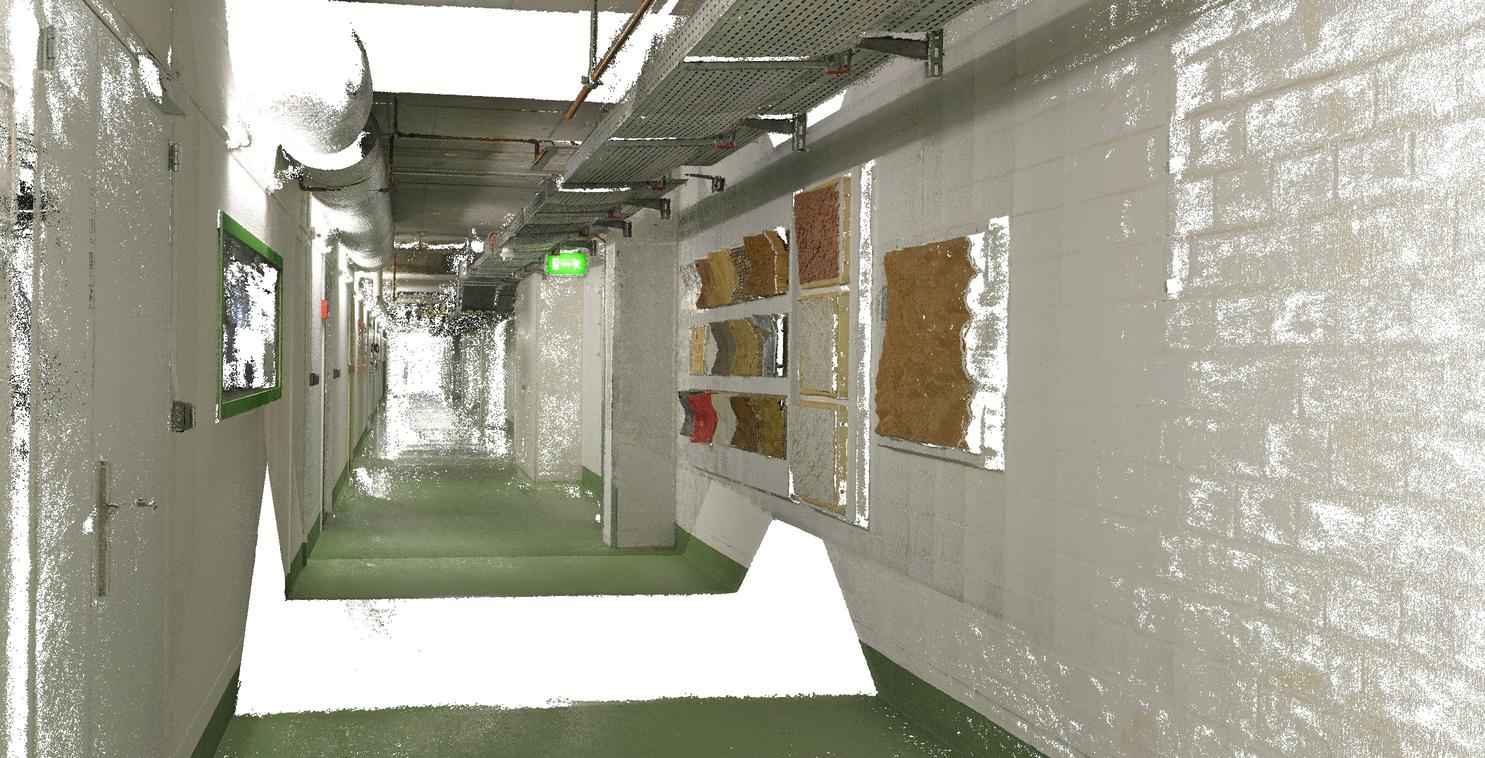}}
 	\vspace{3pt}
 \end{minipage}

\caption{Point cloud results of the ETH3D training dataset.}
\label{fig11}

\end{figure*}

\begin{figure*}[tb]
 \centering

 \begin{minipage}{.1\textwidth}
    \centerline{Botanical}
    \centerline{garden}
    \centerline{\&}
    \centerline{Boulders}
 \end{minipage}%
 \begin{minipage}{.35\linewidth}
 	\vspace{3pt}
 	\centerline{\includegraphics[width=\textwidth]{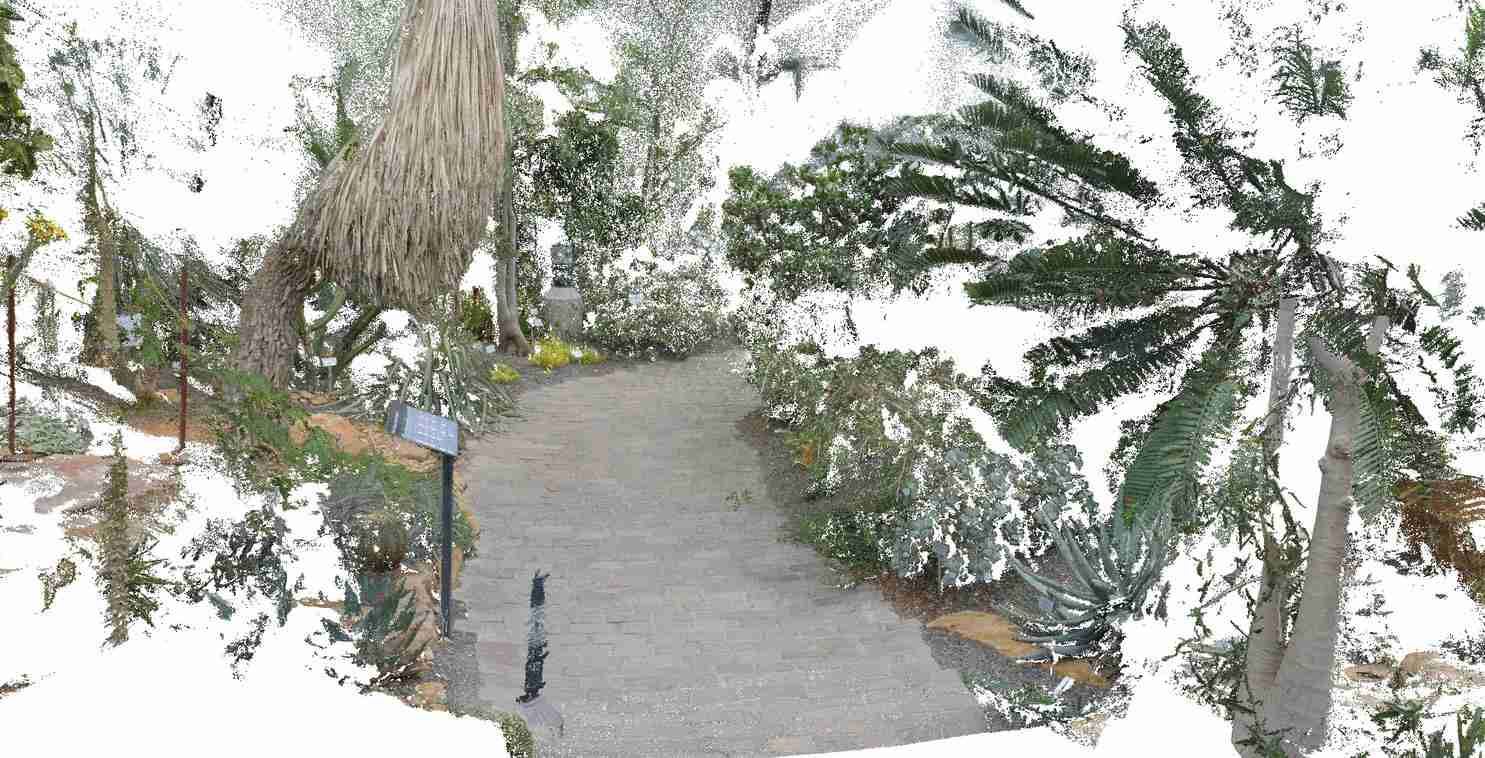}}
 	\vspace{3pt}
 \end{minipage}
  \begin{minipage}{.35\linewidth}
 	\vspace{3pt}
 	\centerline{\includegraphics[width=\textwidth]{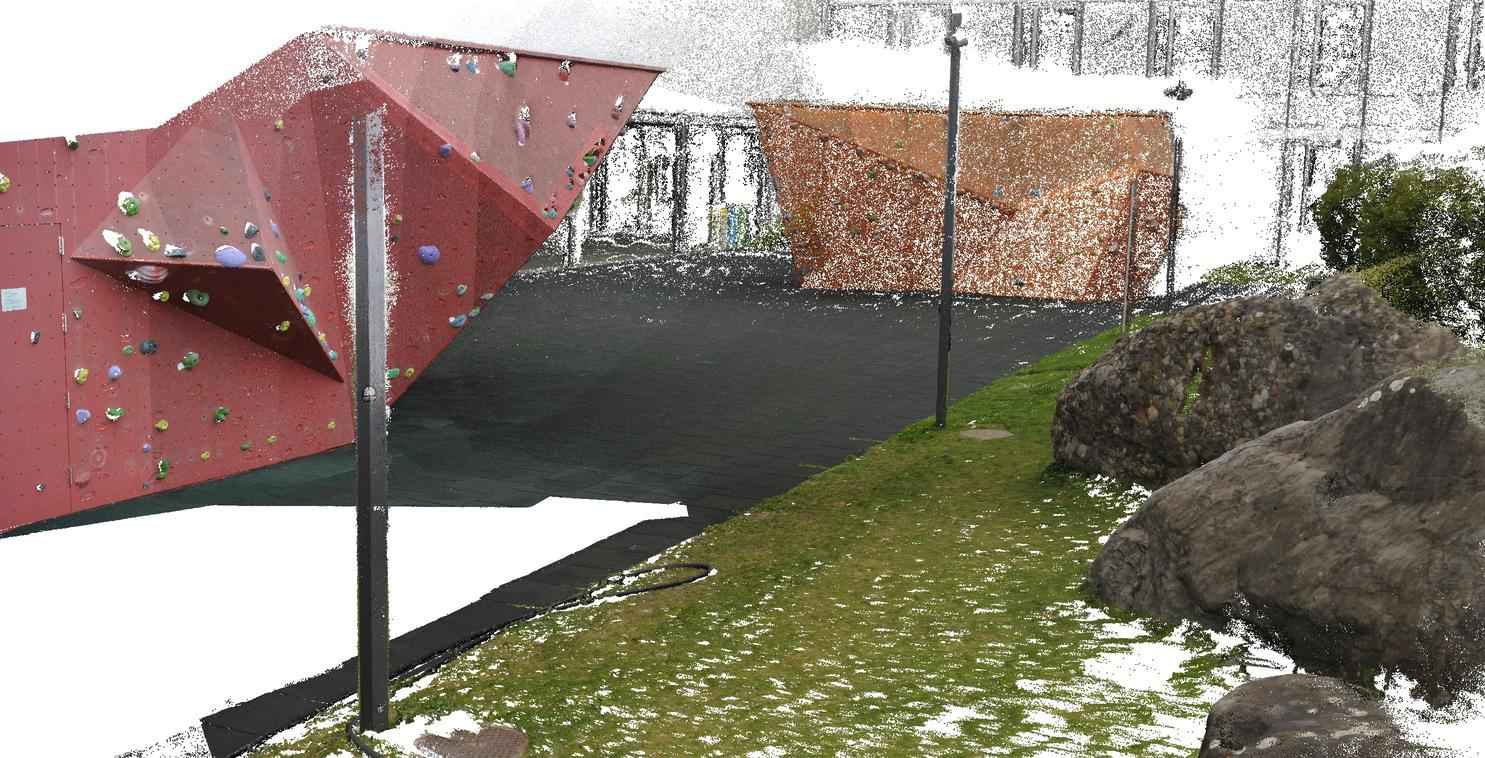}}
 	\vspace{3pt}
 \end{minipage}
 
  \begin{minipage}{.1\textwidth}
    \centerline{Bridge}
    \centerline{\&}
    \centerline{Exhibition}
    \centerline{hall}
 \end{minipage}%
 \begin{minipage}{.35\linewidth}
 	\vspace{3pt}
 	\centerline{\includegraphics[width=\textwidth]{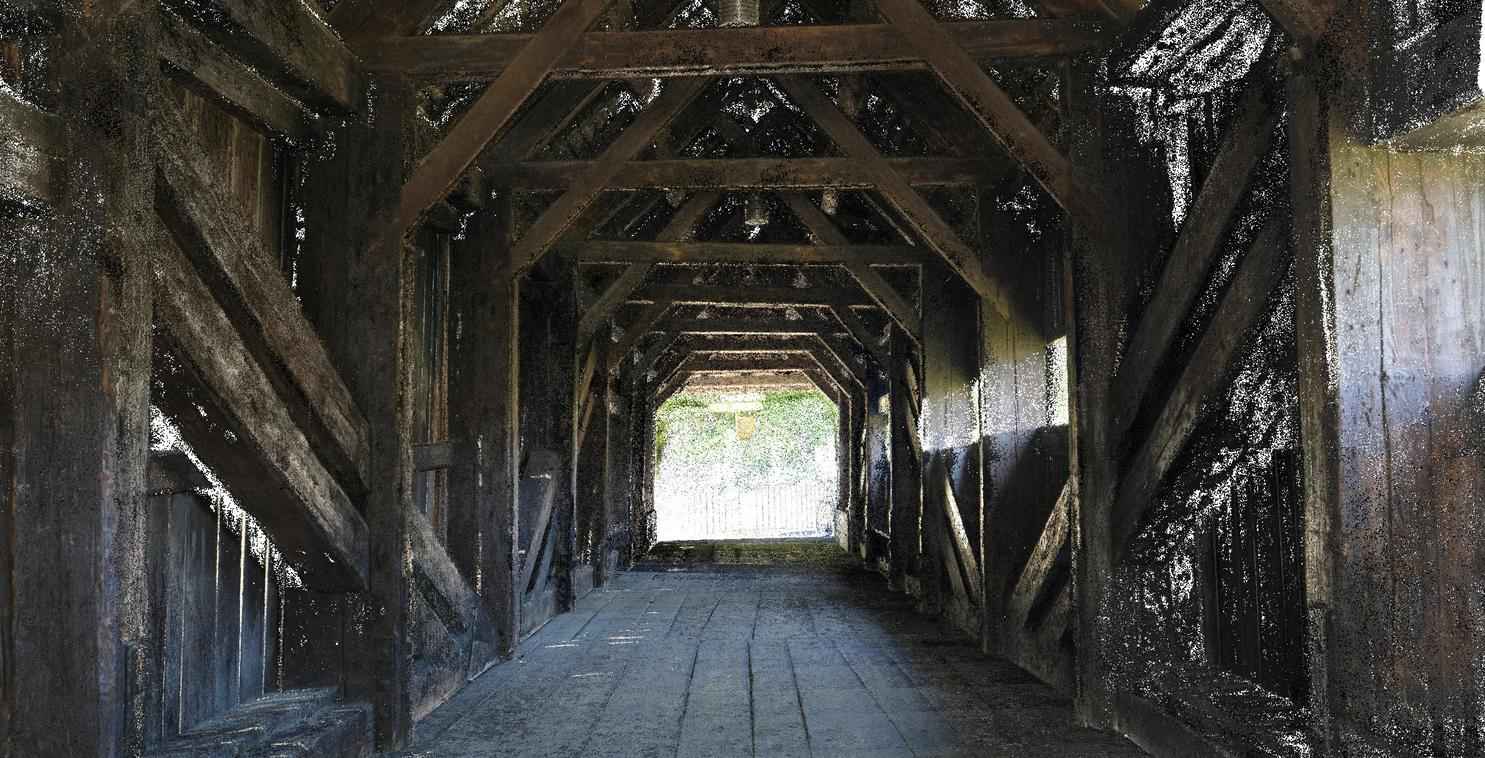}}
 	\vspace{3pt}
 \end{minipage}
  \begin{minipage}{.35\linewidth}
 	\vspace{3pt}
 	\centerline{\includegraphics[width=\textwidth]{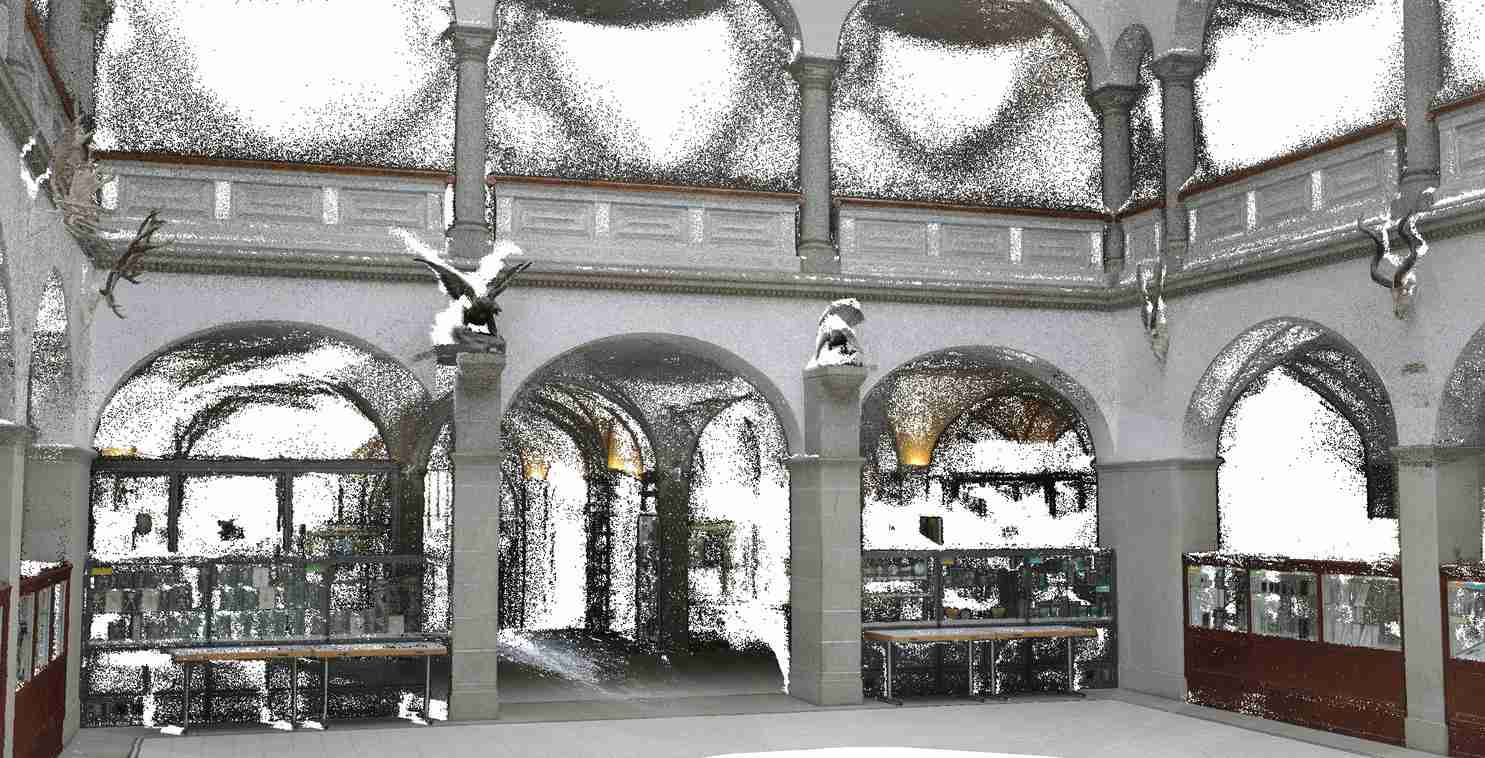}}
 	\vspace{3pt}
 \end{minipage}
 
 \begin{minipage}{.1\textwidth}
    \centerline{Lecture}
    \centerline{room}
    \centerline{\&}
    \centerline{Living}
    \centerline{room}
 \end{minipage}%
 \begin{minipage}{.35\linewidth}
 	\vspace{3pt}
 	\centerline{\includegraphics[width=\textwidth]{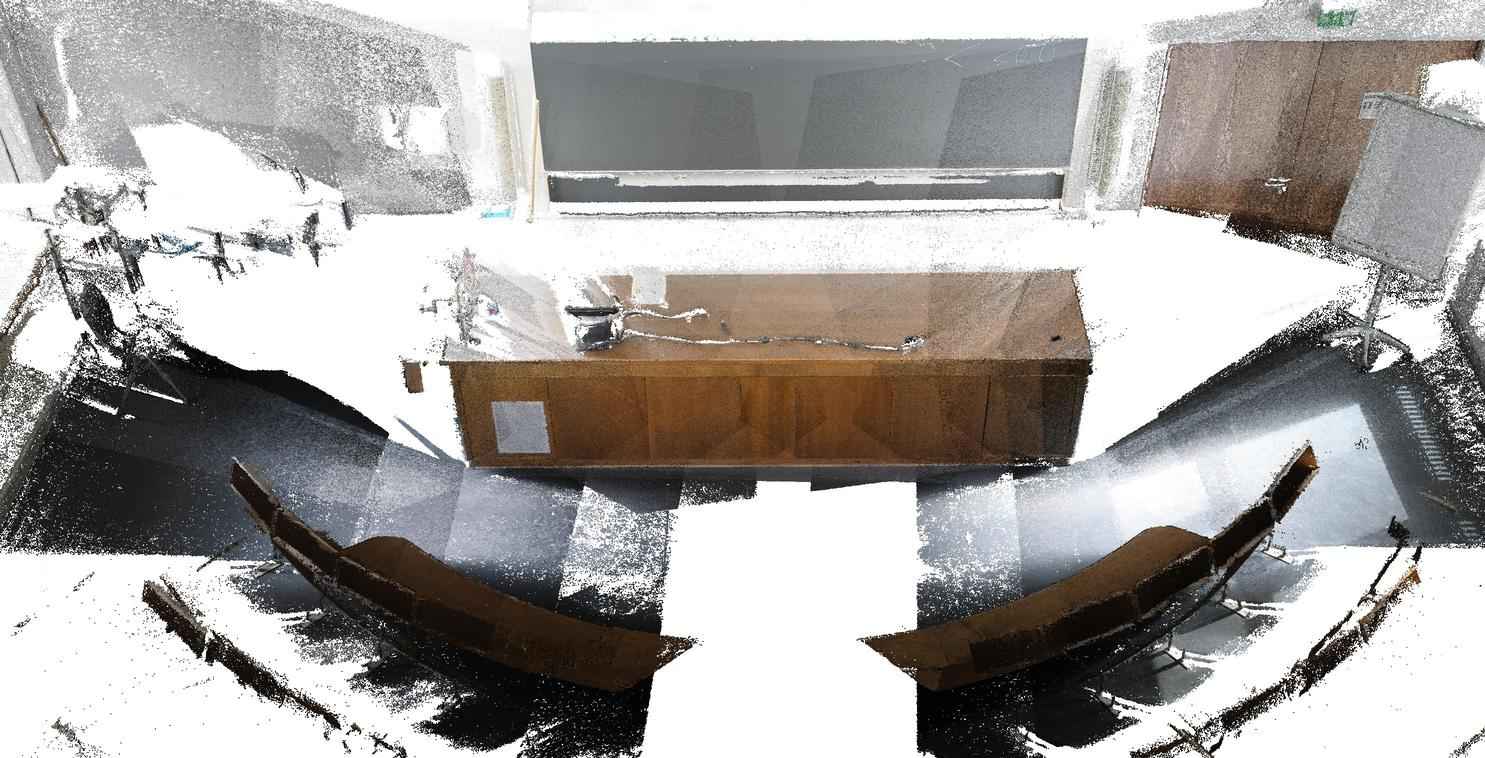}}
 	\vspace{3pt}
 \end{minipage}
  \begin{minipage}{.35\linewidth}
 	\vspace{3pt}
 	\centerline{\includegraphics[width=\textwidth]{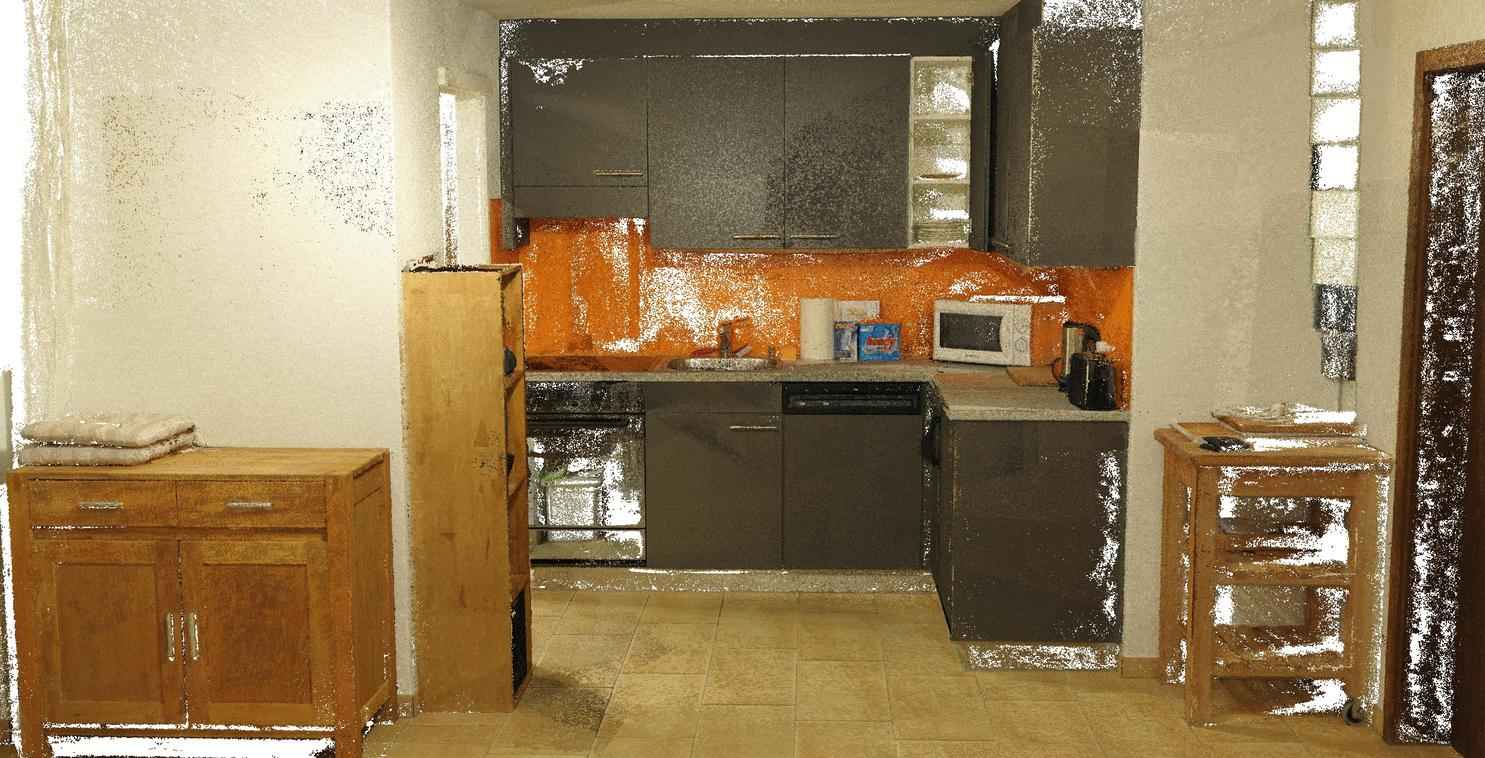}}
 	\vspace{3pt}
 \end{minipage}
 
  \begin{minipage}{.1\textwidth}
    \centerline{Lounge}
    \centerline{\&}
    \centerline{Observatory}
 \end{minipage}%
 \begin{minipage}{.35\linewidth}
 	\vspace{3pt}
 	\centerline{\includegraphics[width=\textwidth]{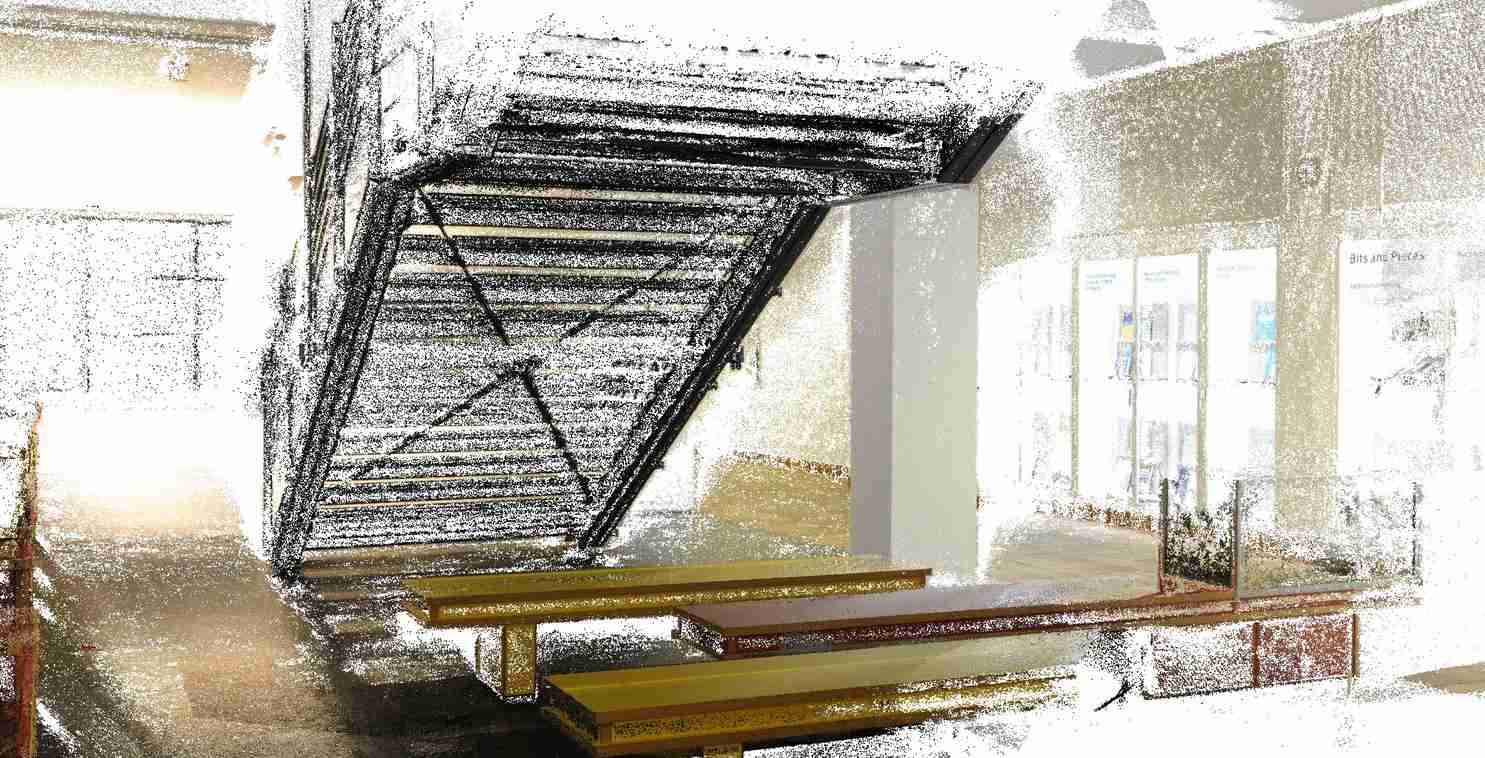}}
 	\vspace{3pt}
 \end{minipage}
  \begin{minipage}{.35\linewidth}
 	\vspace{3pt}
 	\centerline{\includegraphics[width=\textwidth]{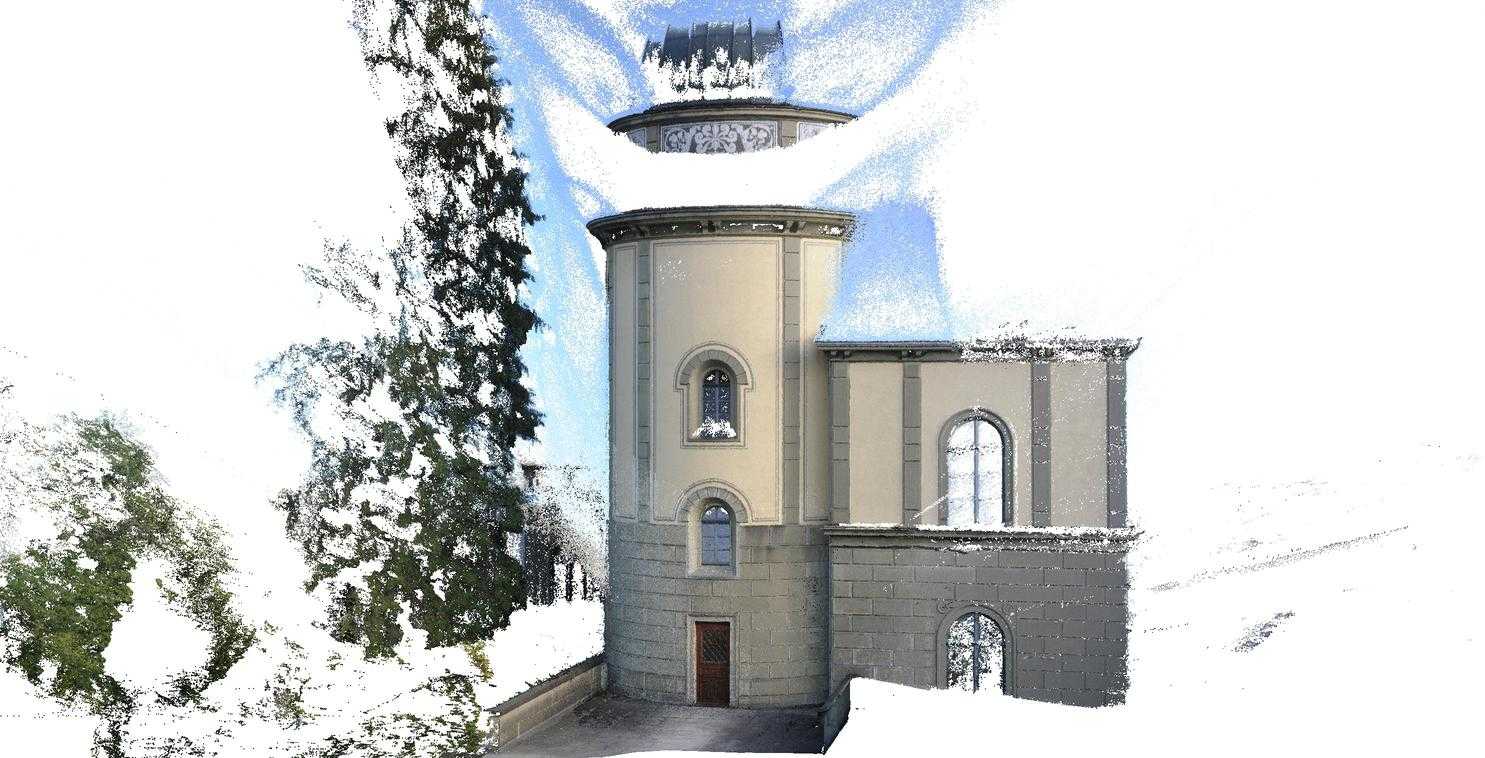}}
 	\vspace{3pt}
 \end{minipage}
 
  \begin{minipage}{.1\textwidth}
    \centerline{Old}
    \centerline{computer}
    \centerline{\&}
    \centerline{Terrace 2}
 \end{minipage}%
 \begin{minipage}{.35\linewidth}
 	\vspace{3pt}
 	\centerline{\includegraphics[width=\textwidth]{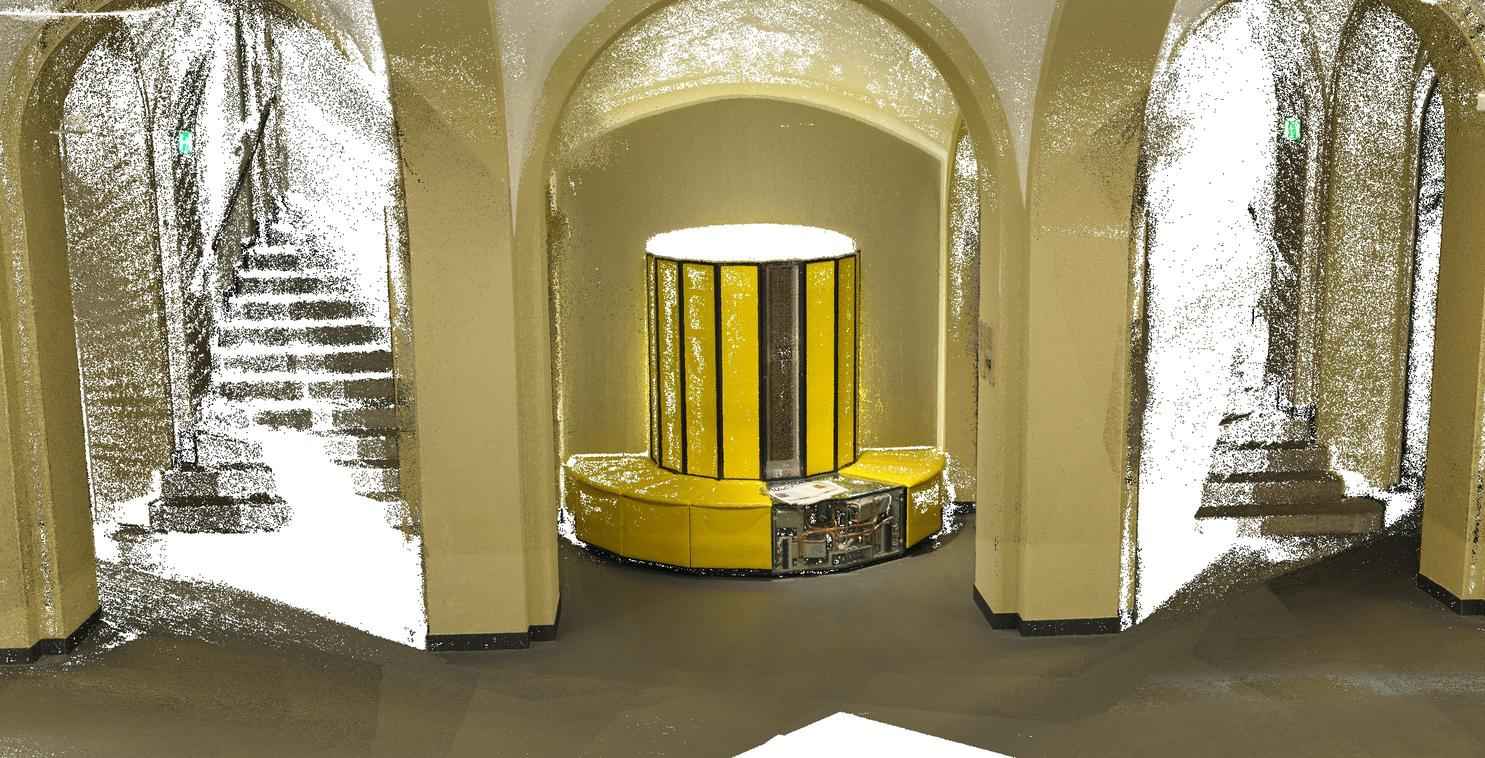}}
 	\vspace{3pt}
 \end{minipage}
  \begin{minipage}{.35\linewidth}
 	\vspace{3pt}
 	\centerline{\includegraphics[width=\textwidth]{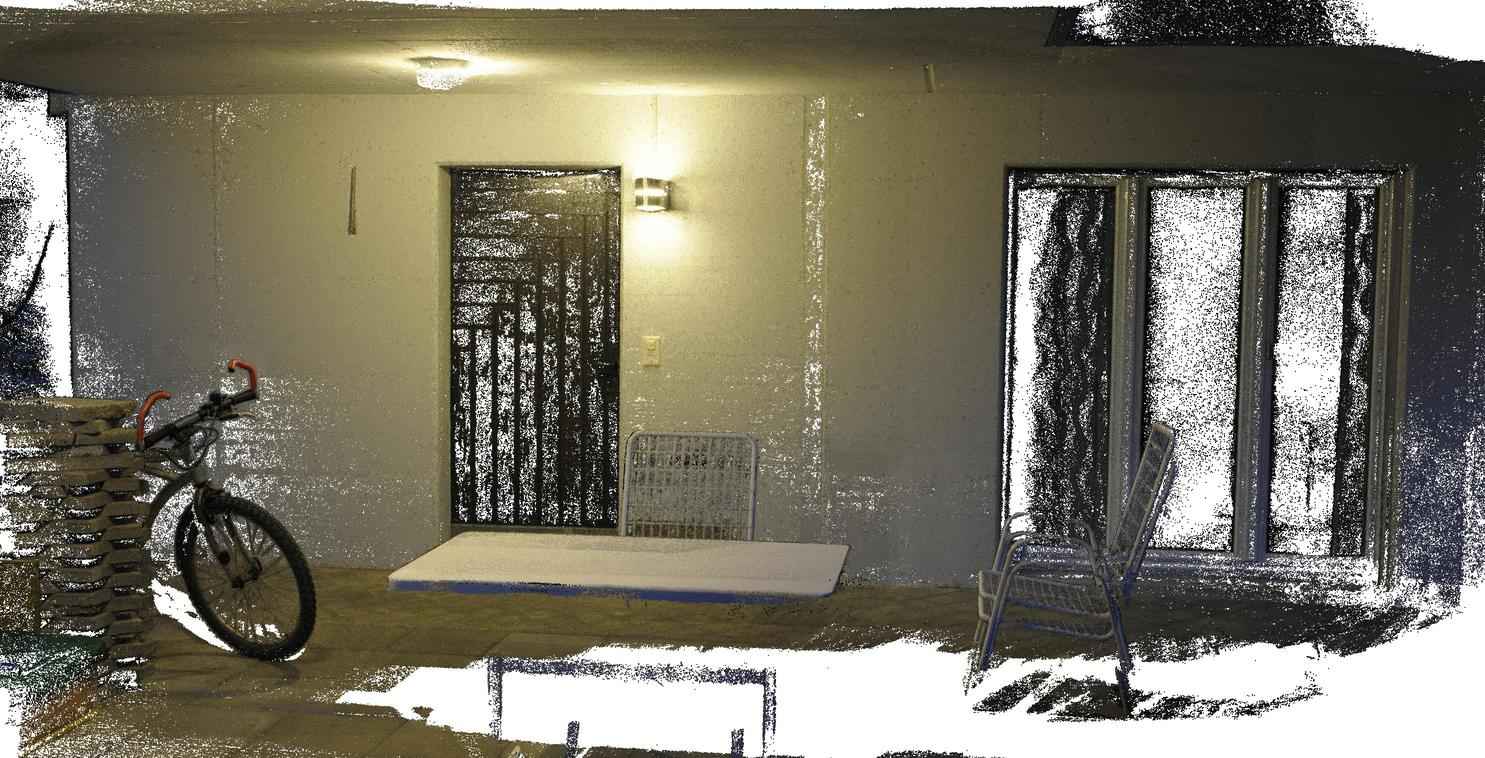}}
 	\vspace{3pt}
 \end{minipage}
 
 \begin{minipage}{.1\textwidth}
    \centerline{Door}
    \centerline{\&}
    \centerline{Statue}
 \end{minipage}%
 \begin{minipage}{.35\linewidth}
 	\vspace{3pt}
 	\centerline{\includegraphics[width=\textwidth]{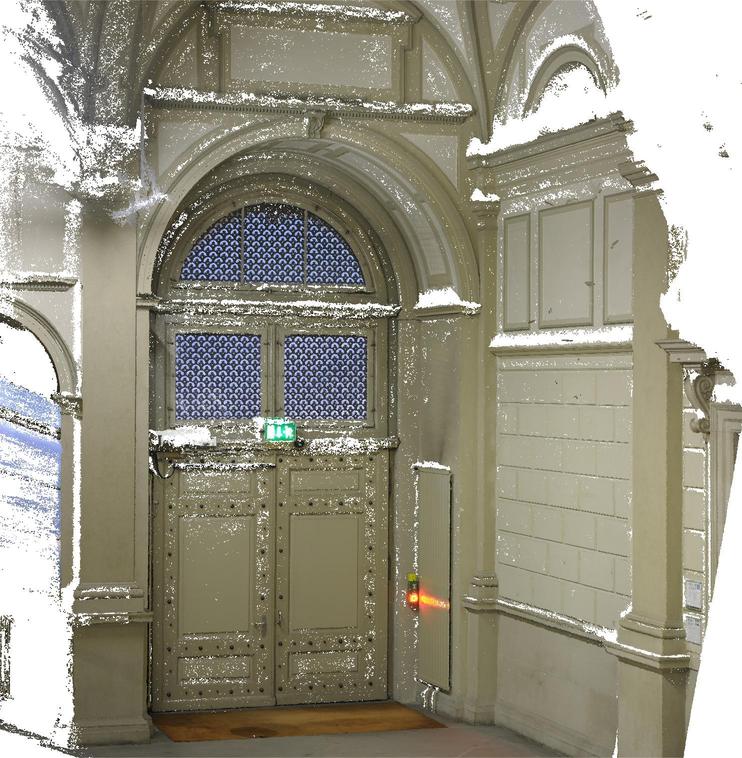}}
 	\vspace{3pt}
 \end{minipage}
 \begin{minipage}{.35\linewidth}
 	\vspace{3pt}
 	\centerline{\includegraphics[width=\textwidth]{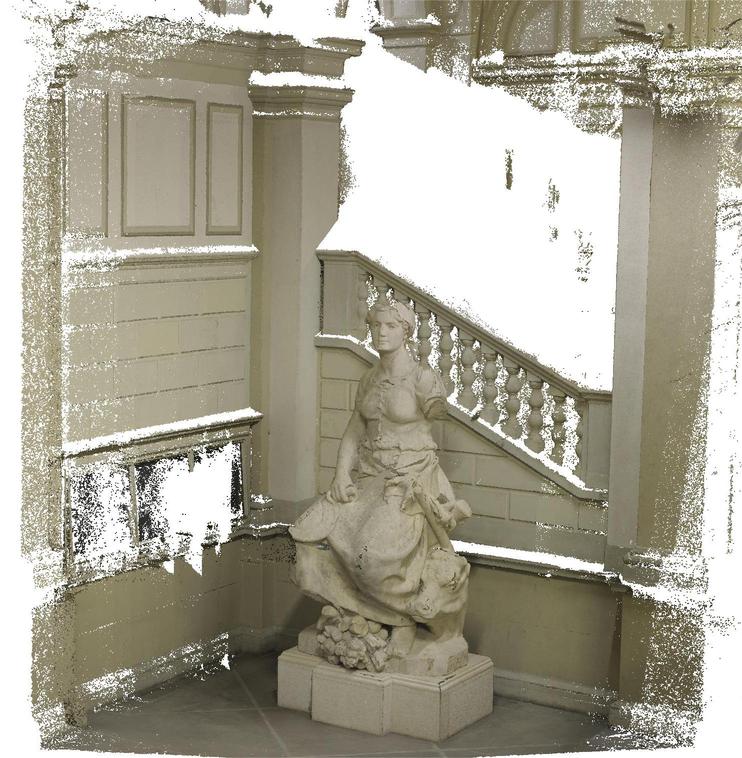}}
 	\vspace{3pt}
 \end{minipage}

\caption{Point cloud results of the ETH3D test dataset.}
\label{fig12}

\end{figure*}

\end{document}